\documentclass[lettersize,journal]{IEEEtran}
\usepackage{cite}

\usepackage{amsmath,amsfonts}
\usepackage{amssymb}
\usepackage{algorithmic}
\usepackage{array}
\usepackage{textcomp}
\usepackage{url}
\usepackage{verbatim}
\usepackage{graphicx}
\usepackage{booktabs}
\usepackage{makecell}
\usepackage{multirow}
\usepackage[table]{xcolor}
\usepackage[numbers,sort&compress]{natbib}

\usepackage[hidelinks]{hyperref}

\def\BibTeX{{\rm B\kern-.05em{\sc i\kern-.025em b}\kern-.08em
    T\kern-.1667em\lower.7ex\hbox{E}\kern-.125emX}}

\begin{document}

\title{UniBuild: Unified Building Mapping From Multi-Source Optical Remote Sensing Imagery With Detail Decoding and Geometry Regularization}

\author{Wei Huang,~\IEEEmembership{Member,~IEEE,}
Chenying Liu,~\IEEEmembership{Member,~IEEE,}
Yilei Shi,~\IEEEmembership{Member,~IEEE,} \\
and Xiao Xiang Zhu,~\IEEEmembership{Fellow,~IEEE}%
\thanks{Wei Huang, Chenying Liu, Yilei Shi, and Xiao Xiang Zhu are with the Chair of Data Science in Earth Observation, Technical University of Munich, 80333 Munich, Germany; Chenying Liu and Xiao Xiang Zhu are also with the Munich Center for Machine Learning, 80333 Munich, Germany.}
\thanks{Corresponding author: Xiao Xiang Zhu.}
\thanks{E-mail: \{w2wei.huang, chenying.liu, yilei.shi, xiaoxiang.zhu\}@tum.de.}
\thanks{Part of the methodological foundation of this work builds upon our paper accepted by NeurIPS 2026.}
}

\markboth{Submitted to IEEE Jounral}%
{Huang \MakeLowercase{\textit{et al.}}: UniBuild: Unified Building Mapping from Open Optical Remote Sensing Imagery With Detail Decoding and Geometry Regularization}

\maketitle

\begin{abstract}
Building extraction from optical remote sensing (RS) imagery is fundamental to urban mapping, yet existing methods are often dataset-specific and generalize poorly to unseen domains. Their practical use is also limited by insufficient detail recovery and weak geometric regularization, leading to blurred boundaries, irregular shapes, and merged adjacent buildings. To address these issues, we propose \textbf{UniBuild}, a unified building extraction framework for multi-source RGB optical RS imagery. First, a unified multi-dataset training scheme is constructed over heterogeneous RGB optical datasets to learn transferable building representations across sensors and resolutions. Second, a novel detail-preserving \textbf{HR-DPT decoder} is designed to integrate high-level semantic features with high-resolution spatial features, enhancing building detail recovery. Third, geometry-aware regularization is introduced through a structure-tensor-based \textbf{direction-aware loss} for boundary direction consistency and a \textbf{saddle-aware loss} for suppressing false activations in narrow inter-building gaps under low-resolution conditions.
We train and evaluate UniBuild on multi-source RGB optical datasets, including 10 public high-resolution datasets and two self-collected low-resolution datasets. Experiments show that UniBuild consistently improves building-region accuracy, boundary sharpness, and adjacent-building separation across diverse datasets. It also generalizes well to unseen domains and supports practical building extraction from RGB optical RS imagery up to 10\,m resolution. The predicted masks can be further converted into GIS-compatible building footprints through simple polygonization. The trained model and inference code are released at \url{https://github.com/zhu-xlab/UniBuild}.

\end{abstract}


\section{Introduction}

Building extraction is a fundamental task in optical remote sensing (RS) image understanding, providing essential spatial information for urban planning, disaster assessment, and map updating. With the rapid accumulation of multi-source optical RS imagery and the development of deep learning, building extraction performance has improved significantly. However, most existing methods are still developed and evaluated on specific datasets, making them difficult to apply to diverse RS imagery in real-world scenarios. Therefore, developing a unified building extraction model for multi-source optical RS imagery is highly desirable. However, achieving this goal remains challenging due to the following key issues.

First, \textbf{cross-sensor and cross-resolution generalization is limited}. Optical RS imagery from different platforms exhibits large variations in spatial resolution, imaging conditions, and appearance distributions, while single public datasets usually cover only limited data diversity. Models trained on such datasets therefore tend to learn data-specific representations and degrade when transferred to unseen sensors or resolutions.

Second, \textbf{HR detail decoding remains insufficient}. Building extraction requires accurate recovery of boundaries, corners, and small structures. Although visual foundation models (VFMs), such as DINOv2 and DINOv3 \citep{oquab2024dinov2,simeoni2025dinov3}, capture strong global semantics, they are still limited in local detail representation and boundary recovery, often producing blurred boundaries and irregular building shapes.

\begin{figure}[h]
    \centering
    \includegraphics[width=1.0\linewidth]{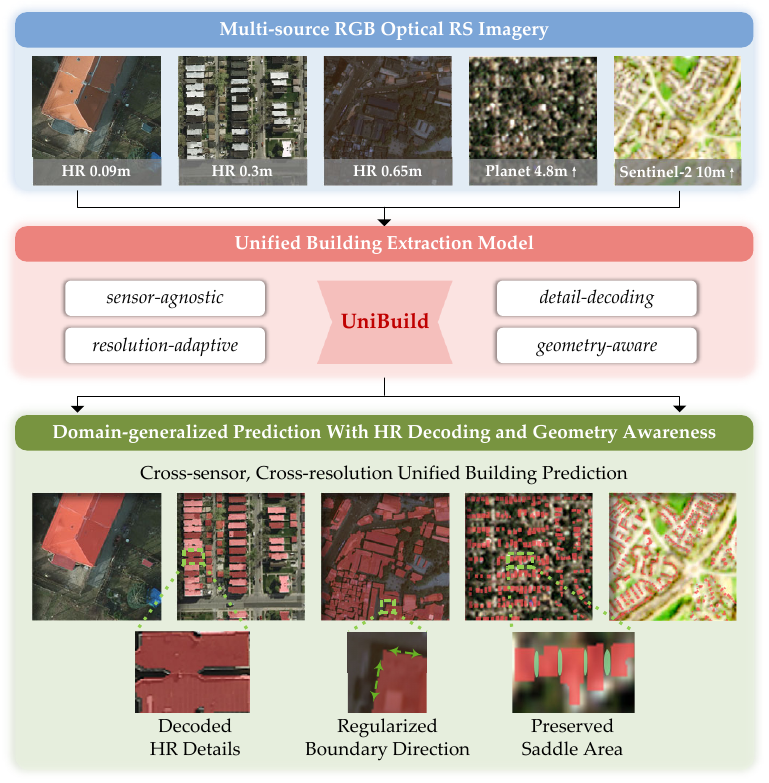}
    \caption{Key properties expected from UniBuild for unified building extraction across multi-source RGB optical RS imagery.}
    \label{fig:goal_overview}
\end{figure}

Third, \textbf{building geometry regularization remains insufficient}. CE and Dice losses mainly optimize pixel-wise accuracy and region overlap, but provide limited constraints on building-specific geometry. On the one hand, weak boundary direction constraints can lead to fragmented, zigzag, or locally inconsistent contours. On the other hand, under LR conditions, mixed pixels and over-smoothed features can obscure narrow inter-building gaps, making these \emph{\textbf{saddle areas}} prone to false foreground activation and causing adjacent buildings to merge. Therefore, unified building extraction requires geometry-aware regularization that jointly improves boundary consistency and preserves LR inter-building gaps.

Based on these observations, we propose \textbf{UniBuild}, a unified building extraction framework for multi-source RGB optical RS imagery, instead of training separate models for different data sources. As shown in Fig.~\ref{fig:goal_overview}, UniBuild is designed to be \emph{\textbf{sensor-agnostic}}, \emph{\textbf{resolution-adaptive}}, \emph{\textbf{detail-decoding}}, and \emph{\textbf{geometry-aware}}. It is jointly trained on multiple public and private datasets covering diverse RGB optical sensors and spatial resolutions. To recover HR details, we design an \textbf{HR-DPT decoder} that fuses high-level semantic features with high-resolution shallow features. To enhance geometric modeling, we further introduce a structure-tensor-based \textbf{direction-aware loss} for boundary direction consistency and a \textbf{saddle-aware loss} for suppressing false activations in narrow inter-building gaps under LR conditions.

We train and evaluate UniBuild on 10 public HR RGB building extraction datasets and two self-collected LR RGB datasets derived from 4.8\,m Planet and 10\,m Sentinel-2 imagery. By integrating HR, MR, and LR imagery into a single framework, UniBuild establishes a practical unified building extraction model for \textbf{RGB optical RS imagery up to 10\,m resolution}. Experimental results show that UniBuild achieves stable improvements across datasets, especially in boundary quality, and demonstrates stronger cross-sensor and cross-resolution generalization than baselines using generic feature extraction models and conventional pixel-level losses.

In addition, we released the trained UniBuild model and inference code, which can be directly applied to arbitrary RGB optical RS imagery without additional training or fine-tuning. For imagery coarser than 1\,m, only a simple upsampling step to 1\,m GSD is required before inference. The pipeline also integrates optional polygonization, enabling users to convert RGB imagery into GIS-ready building footprint shapefiles for downstream geospatial applications.

In summary, the contributions of this work are as follows:
\begin{itemize}
    \item We propose \textbf{UniBuild}, a unified building extraction framework for optical RS imagery that produces building masks and optional vectorized footprints. UniBuild achieves robust cross-sensor and cross-resolution generalization for \textbf{RGB optical imagery up to 10\,m}.
    
    \item We design a novel \textbf{HR-DPT decoder} that combines high-level semantic features with high-resolution shallow details to improve HR detail decoding.
    
    \item We introduce a novel \textbf{direction-aware loss} that regularizes local boundary direction consistency across all resolutions and strengthens geometric relationship modeling.
    
    \item We propose a novel \textbf{saddle-aware loss} to suppress false activations in LR inter-building gaps and improve adjacent-building separability.
\end{itemize}
\section{Related Work}

\subsection{Building Extraction from Optical Remote Sensing Imagery}

Building extraction from optical RS imagery is a core geospatial vision task for urban monitoring, disaster assessment, map updating, and infrastructure analysis \citep{luo2021deep}. Recent reviews further summarize building extraction from geometrical and semantic perspectives \citep{li2024reviewbuilding}, while large-scale open building datasets highlight the need for globally complete and structurally rich building representations \citep{zhu2025globalbuildingatlas}. Early automatic methods mainly relied on handcrafted spectral, texture, geometric, and height cues, often combined with rule-based reasoning or probabilistic graphical models \citep{awrangjeb2010automatic,li2015robust}.

Deep learning has advanced building extraction into end-to-end semantic segmentation.
CNN-based encoder--decoder architectures such as FCN \citep{long2015fully}, U-Net \citep{ronneberger2015unet}, and SegNet \citep{badrinarayanan2017segnet} established the foundation for many RS building extraction pipelines. FCN-style RS methods demonstrated strong potential for building delineation \citep{bittner2017building,ji2019multisource}, and related studies further explored joint road--building extraction \citep{alshehhi2017simultaneous}, multimodal HR image--LiDAR fusion \citep{li2019grrnet}, cross-source adaptive fusion \citep{song2025mcfnet,song2026multi}, and graph-based structural refinement \citep{shi2020ggcn}. Subsequent works improved building extraction through guided filtering, attention mechanisms, body--boundary decomposition, and multimodal feature fusion \citep{xu2018building,li2024hdnet,li2019grrnet,huang2025heightmatch}.

More recently, transformer-based methods and VFMs have brought new opportunities by providing stronger long-range dependency modeling and transferable representations. Sparse Token Transformers demonstrated the effectiveness of transformers for RS building extraction \citep{chen2021building}, while large-scale self-supervised VFMs such as DINOv2 \citep{oquab2024dinov2} and DINOv3 \citep{simeoni2025dinov3} showed strong transferability across domains and dense prediction tasks. Building extraction has also been investigated under self-supervised, semi-supervised, weakly supervised, unsupervised domain adaptation, and vectorized settings \citep{zhu2024unrestricted,huang2023adaptmatch,huang2024dbmatch,huang2025heightmatch,liu2024aio2,chen2023mcuda,du2024vectorized}. However, most existing methods are still developed and validated on one or only a few datasets, making them insufficient for a unified setting where one shared model must generalize across heterogeneous RGB optical RS imagery with different sensors and spatial resolutions.

\subsection{High-Resolution Detail Decoding for Dense Prediction}

HR detail recovery remains a key challenge in dense prediction, as encoder downsampling weakens spatial cues for boundary reconstruction. Classical encoder--decoder architectures alleviate this issue through skip connections and multi-stage upsampling \citep{long2015fully,ronneberger2015unet,badrinarayanan2017segnet}, while multi-scale designs improve contextual aggregation and feature fusion \citep{lin2017feature,zhao2017pyramid}. In RS imagery, semantic--spatial refinement has also proven effective for geometry-sensitive tasks such as road extraction \citep{yang2026semantic}.

With transformer-based dense prediction, decoder design becomes crucial for converting semantically strong but spatially compressed features into fine-grained outputs. SegFormer supports efficient dense prediction with hierarchical transformer features \citep{xie2021segformer}, while DPT reassembles transformer features into multi-scale image-like representations for progressive refinement \citep{ranftl2021dpt}. These studies show that dense prediction quality depends heavily on how high-level semantics are decoded into HR spatial predictions.

This issue is especially important for building extraction, where sharp corners and thin boundaries must be preserved. Although VFMs such as DINOv2 \citep{oquab2024dinov2} and DINOv3 \citep{simeoni2025dinov3} provide strong transferable semantics, their predictions are still decoded from spatially compressed embeddings, which may lead to blurred contours, incomplete structures, and missing small buildings, particularly in LR imagery. Therefore, unified building extraction requires a decoder that combines high-level semantic embeddings with stronger HR features for boundary recovery and local detail restoration.

\vspace{-0.2cm}

\subsection{Geometry-Aware Regularization for Building Extraction}

Geometry-aware regularization has gained attention for constraining object structures, especially in building extraction, where predictions require regular boundaries, plausible shapes, and clear separation between adjacent buildings. Existing methods can be broadly grouped into boundary-aware and structure-/topology-aware regularization.

Boundary-aware methods explicitly encourage better alignment between predicted and reference contours. Representative examples include Boundary Loss \citep{kervadec2019boundary} and Active Boundary Loss \citep{wang2022active}. In optical RS building extraction, boundary-assisted learning \citep{he2021boundary}, conditional random fields \citep{li2015robust,zhu2020building}, dual spatial-graph refinement \citep{deng2023dual}, and edge-aware refinement networks such as BEARNet \citep{lin2023bearnet} have further demonstrated the benefit of boundary modeling for improving building morphology. However, these methods mainly emphasize contour sharpness or boundary alignment, while direction consistency along building boundaries is less explicitly modeled.

Structure- and topology-aware methods aim to encode higher-level geometric priors into segmentation. For example, clDice \citep{shit2021cldice} preserves connectivity by introducing topology-aware supervision. In building extraction, adversarial shape learning \citep{ding2022aslnet} and vectorized building outline modeling \citep{du2024vectorized} further highlight the importance of regular shapes and structurally meaningful predictions. These studies show that building extraction requires supervision beyond independent pixels, as buildings should exhibit coherent edges, plausible geometry, and clear separation from neighboring objects.

Despite these advances, existing methods still insufficiently model local directions along building boundaries and pay limited attention to narrow inter-building gaps, especially under LR conditions where mixed pixels and over-smoothed representations can merge adjacent buildings. More broadly, robust cross-sensor and cross-resolution generalization, HR detail decoding, and geometry-aware regularization for boundary consistency and gap preservation remain underexplored. These limitations motivate UniBuild, a unified framework with resolution-adaptive representation, detail-preserving decoding, and complementary geometry-aware supervision.

\section{UniBuild}

\begin{figure*}[h]
    \centering
    \includegraphics[width=1.0\linewidth]{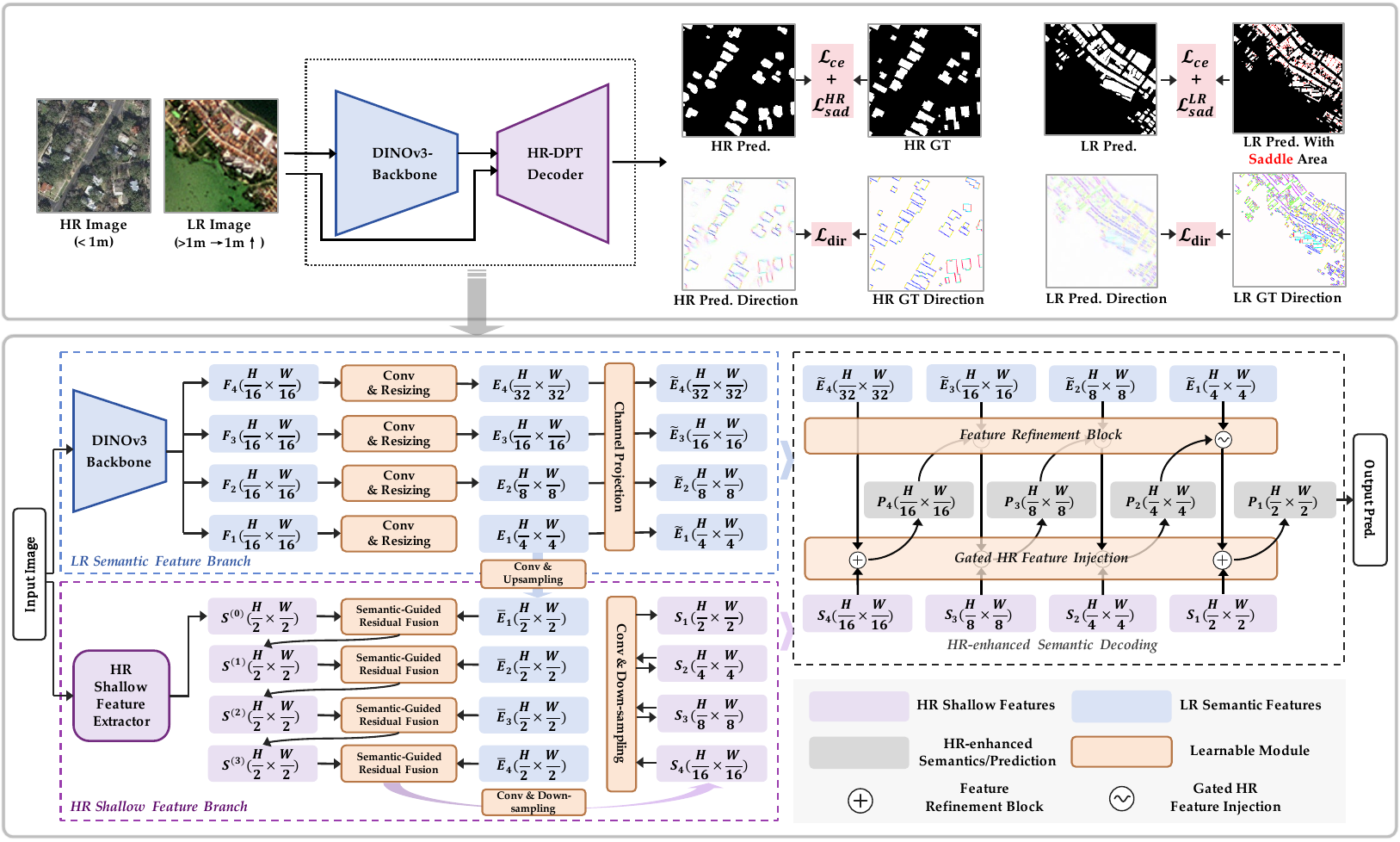}
    \caption{Overview of UniBuild. Multi-source RGB optical RS images are processed by a DINOv3 backbone and the proposed HR-DPT decoder, while direction-aware and saddle-aware losses regularize boundary orientation and LR inter-building gaps.}
    \label{fig:unibuild-framework}
\end{figure*}

This section presents UniBuild from four aspects: unified input representation and VFM feature extraction, the proposed HR-DPT decoder, direction-aware boundary regularization, and saddle-aware LR gap suppression, as shown in Fig.~\ref{fig:unibuild-framework}.

\subsection{Unified Input Representation and VFM Feature Extraction}

UniBuild aims to learn a shared building extraction model from multi-source RGB optical RS imagery with heterogeneous sensors and spatial resolutions. To make such inputs compatible, we first construct a unified input representation.

\subsubsection{Resolution-aware input normalization and batch grouping}

Input images are divided into HR and LR groups using 1,m GSD as the threshold, which balances building-detail preservation and computational cost. HR images are processed at their original resolution, while LR images are bilinearly upsampled to 1\,m GSD. All images are then cropped into $512\times512$ patches. During training, each mini-batch contains samples from only one resolution group, enabling resolution-specific optimization. Specifically, saddle-aware regularization is activated only for LR batches, while direction-aware regularization is applied to both HR and LR batches.

\subsubsection{Unified augmentation and binary supervision}

Let $\mathbf{I}\in\mathbb{R}^{3\times H\times W}$ denote the input RGB image and $\mathbf{Y}\in{0,1}^{H\times W}$ the binary building mask. During training, we apply weak geometric augmentation and strong intensity augmentation. Although the datasets differ in resolution, annotation quality, and label taxonomy, all annotations are converted into a unified binary label space of building and background.

\subsubsection{VFM-based multi-level feature extraction}

Given an input image $\mathbf{I}$, DINOv3 is adopted as the VFM backbone $\mathcal{B}(\cdot)$. The image is tokenized and processed to obtain four intermediate feature maps:
\begin{equation}
\label{eq:vfm-feature-extraction}
\{\mathbf{F}_1,\mathbf{F}_2,\mathbf{F}_3,\mathbf{F}_4\} = \mathcal{B}(\mathbf{I}).
\end{equation}
These non-hierarchical DINOv3 features share the same spatial resolution of $H/16\times W/16$ while encoding multi-depth semantic cues. However, such spatial compression limits the recovery of fine building boundaries, corners, and small structures, motivating the proposed HR-DPT decoder.

\subsection{HR-DPT Decoder}

The VFM backbone provides strong semantic representations but limited spatial details. To recover HR building details, we propose an \textbf{HR-DPT decoder} with an LR semantic branch and a semantics-guided HR shallow branch. The LR branch performs DPT-style top-down decoding \citep{ranftl2021dpt}, while the HR branch preserves fine image details and injects them into the semantic stream through gated residual fusion.

\subsubsection{LR semantic feature branch}

Given the backbone features
\begin{equation}
\label{eq:backbone-feature-shape}
\mathbf{F}_t \in \mathbb{R}^{C_b\times H/16\times W/16}, \quad t\in\{1,2,3,4\},
\end{equation}
we use $1\times1$ projection and scale-specific resizing to construct a semantic pyramid:
\begin{equation}
\label{eq:semantic-pyramid-construction}
\mathbf{E}_t = \mathcal{R}_t\left(\mathrm{Conv}_t^{1\times1}(\mathbf{F}_t)\right), \quad t\in\{1,2,3,4\}.
\end{equation}
Here, $\mathrm{Conv}_t^{1\times1}(\cdot)$ maps $C_b$ to $C_t$, and $\mathcal{R}_t(\cdot)$ aligns each feature to its pyramid scale:
\begin{equation}
\label{eq:semantic-pyramid-scales}
\begin{gathered}
\mathbf{E}_1 \in \mathbb{R}^{C_1\times H/4\times W/4}, \ 
\mathbf{E}_2 \in \mathbb{R}^{C_2\times H/8\times W/8},\\
\mathbf{E}_3 \in \mathbb{R}^{C_3\times H/16\times W/16},\
\mathbf{E}_4 \in \mathbb{R}^{C_4\times H/32\times W/32}.
\end{gathered}
\end{equation}
The four resizing operations use stride-$4$ and stride-$2$ transposed convolutions, identity mapping, and stride-$2$ convolution, respectively. However, these semantic features are still derived from spatially compressed VFM embeddings, so an additional HR shallow branch is introduced to recover fine spatial details.

\subsubsection{HR shallow feature branch}

An HR shallow feature is first extracted from the input image:
\begin{equation}
\label{eq:initial-hr-shallow-feature}
\mathbf{S}^{(0)} = \psi(\mathbf{I}), \quad \mathbf{S}^{(0)} \in \mathbb{R}^{C_s\times H/2\times W/2},
\end{equation}
where $\psi(\cdot)$ is a stride-$2$ convolution followed by a refinement convolution. For each semantic level $\mathbf{E}_t$, we project and upsample it to the HR feature resolution:
\begin{equation}
\label{eq:aligned-semantic-feature}
\bar{\mathbf{E}}_t = \left(\mathrm{Conv}^{1\times1}_t(\mathbf{E}_t)\right)^{\uparrow}, \quad t\in\{1,2,3,4\},
\end{equation}
where $(\cdot)^{\uparrow}$ denotes bilinear upsampling to $H/2\times W/2$. The semantic feature is then fused with the HR feature as:
\begin{equation}
\label{eq:semantics-guided-hr-fusion}
\mathbf{S}^{(t)} = \mathbf{S}^{(t-1)} + f_t\left([\mathbf{S}^{(t-1)},\bar{\mathbf{E}}_t]\right), \quad t\in\{1,2,3,4\},
\end{equation}
where $[\cdot,\cdot]$ denotes concatenation and $f_t(\cdot)$ consists of two $3\times3$ convolutional layers. This progressively injects semantic cues into HR features while preserving fine spatial structures.

From $\mathbf{S}^{(4)}$, we build an HR feature pyramid:
\begin{equation}
\label{eq:hr-feature-pyramid-construction}
\begin{gathered}
\mathbf{S}_1 = \mathbf{S}^{(4)}, \ 
\mathbf{S}_2 = \mathcal{D}_2(\mathbf{S}_1),\\
\mathbf{S}_3 = \mathcal{D}_3(\mathbf{S}_2),\
\mathbf{S}_4 = \mathcal{D}_4(\mathbf{S}_3).
\end{gathered}
\end{equation}
Here, $\mathcal{D}_2(\cdot)$, $\mathcal{D}_3(\cdot)$, and $\mathcal{D}_4(\cdot)$ are stride-$2$ downsampling blocks. The resulting scales are
\begin{equation}
\label{eq:hr-feature-pyramid-scales}
\begin{gathered}
\mathbf{S}_1 \in \mathbb{R}^{C_s\times H/2\times W/2},\ 
\mathbf{S}_2 \in \mathbb{R}^{C_s\times H/4\times W/4},\\
\mathbf{S}_3 \in \mathbb{R}^{C_s\times H/8\times W/8},\ 
\mathbf{S}_4 \in \mathbb{R}^{C_s\times H/16\times W/16}.
\end{gathered}
\end{equation}

Before top-down decoding, each semantic feature is projected into the unified decoder space by a linear layer $\rho_t$:
\begin{equation}
\label{eq:semantic-feature-projection}
\tilde{\mathbf{E}}_t = \rho_t(\mathbf{E}_t), \quad t\in\{1,2,3,4\}.
\end{equation}
The projected semantic features have the following scales:
\begin{equation}
\label{eq:projected-semantic-feature-scales}
\begin{gathered}
\tilde{\mathbf{E}}_1 \in \mathbb{R}^{C\times H/4\times W/4},\
\tilde{\mathbf{E}}_2 \in \mathbb{R}^{C\times H/8\times W/8},\\
\tilde{\mathbf{E}}_3 \in \mathbb{R}^{C\times H/16\times W/16},\
\tilde{\mathbf{E}}_4 \in \mathbb{R}^{C\times H/32\times W/32}.
\end{gathered}
\end{equation}
Here, $C$ is the unified decoder channel dimension.

\subsubsection{HR-enhanced semantic decoding}

The semantic branch is decoded in a DPT-style top-down manner. Starting from the coarsest feature, each refinement block upsamples the decoded feature and fuses it with a finer semantic feature. To compensate for missing HR details, the matched HR shallow feature $\mathbf{S}_t$ is injected after each refinement step.

Let $\bar{\mathbf{P}}_t$ and $\mathbf{P}_t$ denote the decoded features before and after HR injection, respectively:
\begin{equation}
\label{eq:semantic-refinement-unified}
\begin{gathered}
\bar{\mathbf{P}}_4 = \mathrm{Refine}_4(\tilde{\mathbf{E}}_4),\\
\bar{\mathbf{P}}_t = \mathrm{Refine}_t(\mathbf{P}_{t+1},\tilde{\mathbf{E}}_t), \quad t\in\{3,2,1\},
\end{gathered}
\end{equation}
where $\mathrm{Refine}_t$ denotes the $t$-th refinement block in DPT.

To control the contribution of HR details, a gate map is calculated from the decoded semantic feature and the matched HR shallow feature:
\begin{equation}
\label{eq:hr-gate-map}
\mathbf{A}_t = \sigma\left(\mathrm{Conv}^{1\times1}_t\left([\bar{\mathbf{P}}_t,\mathbf{S}_t]\right)\right), \quad t\in\{4,3,2,1\}.
\end{equation}
The gated residual HR injection is formulated as
\begin{equation}
\label{eq:gated-hr-feature-injection}
\mathbf{P}_t = \bar{\mathbf{P}}_t + (1+\mathbf{A}_t)\odot\phi_t(\mathbf{S}_t), \quad t\in\{4,3,2,1\}.
\end{equation}
Here, $\phi_t(\cdot)$ projects $\mathbf{S}_t$ to the decoder dimension $C$. This module supplements semantic features with spatially HR details.

Finally, the finest decoded feature $\mathbf{P}_1$ is passed to a prediction head $\mathcal{H}$ and upsampled to the original resolution:
\begin{equation}
\label{eq:final-prediction-map}
\mathbf{P} = \mathrm{Softmax}\left(\left(\mathcal{H}(\mathbf{P}_1)\right)^{\uparrow}\right), \quad \mathbf{P} \in [0,1]^{2\times H\times W},
\end{equation}
where the two classes are background and building.

\subsection{Direction-aware Consistency}

\begin{figure}[h]
    \centering
    \includegraphics[width=1.0\linewidth]{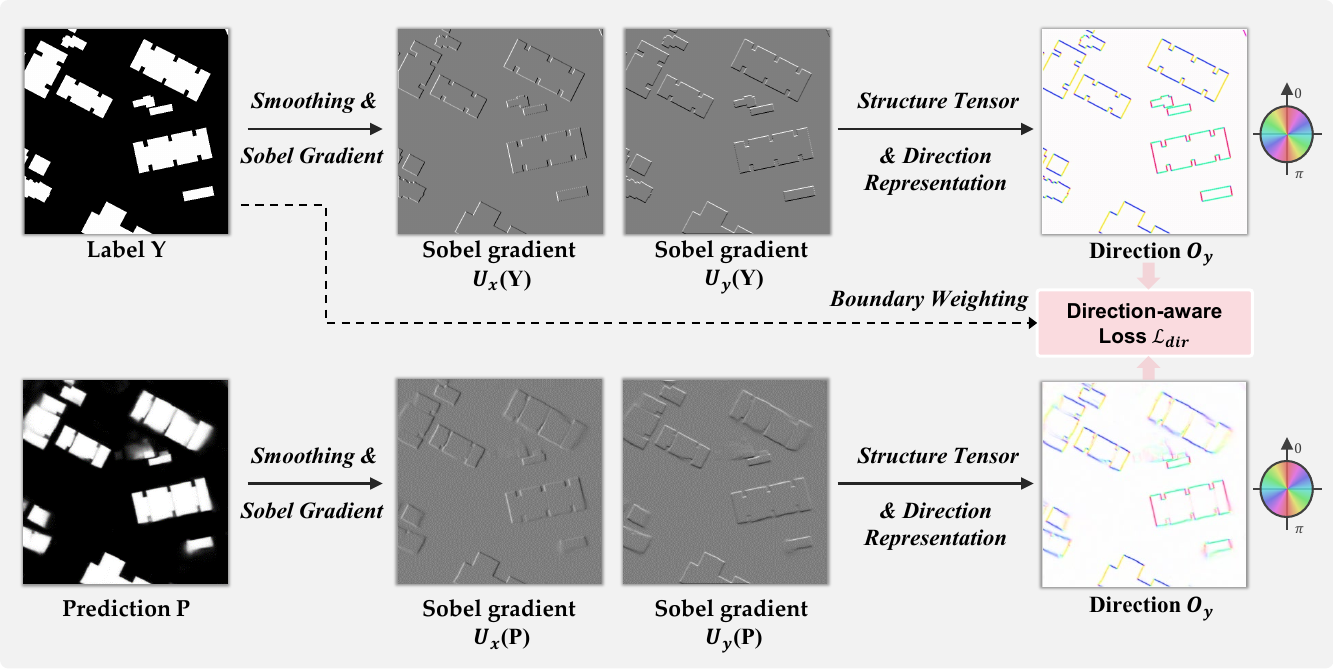}
    \caption{Visualization of the proposed direction-aware loss.}
    \label{fig:direction_aware_loss}
    \vspace{-0.3cm}
\end{figure}

As shown in Fig.~\ref{fig:direction_aware_loss}, to improve boundary regularity, we introduce a \textbf{direction-aware loss} that aligns structure-tensor-based orientation between the predictions and labels, which is imposed only around  boundary-support regions.

\subsubsection{Boundary-focused supervision region}

Let $\mathbf{Y}\in\{0,1\}^{H\times W}$ be the ground-truth label map, with ignored pixels excluded. We convert it into a two-channel one-hot map:
\begin{equation}
\label{eq:one-hot-label-map}
\mathbf{Y}^{\mathrm{oh}}\in\{0,1\}^{2\times H\times W}.
\end{equation}
Since directional structure is meaningful mainly around class transitions, a boundary-support mask $\mathbf{B}$ is constructed by detecting local label variation:
\begin{equation}
\label{eq:boundary-support-mask}
\mathbf{B}
=
\max_{c\in\{0,1\}}
\mathbb{I}
\left(
\mathrm{MaxP}^{3\times3}(\mathbf{Y}^{\mathrm{oh}}_c)
-
\mathrm{MinP}^{3\times3}(\mathbf{Y}^{\mathrm{oh}}_c)
>0
\right).
\end{equation}

\subsubsection{Structure-tensor-based direction representation}

For the prediction branch, we use the probability map $\mathbf{P}\in[0,1]^{2\times H\times W}$. For the target branch, the one-hot label map is smoothed by a $3\times3$ average filter:
\begin{equation}
\label{eq:smoothed-target-map}
\mathbf{Y}^{\mathrm{sm}} = \mathrm{AvgPool}^{3\times3}\left(\mathbf{Y}^{\mathrm{oh}}\right).
\end{equation}
Let $\mathbf{U}$ denote either $\mathbf{P}$ or $\mathbf{Y}^{\mathrm{sm}}$. Class-wise Sobel gradients are computed as
\begin{equation}
\label{eq:sobel-gradients}
U_{x,c}=K_x * U_c, \quad U_{y,c}=K_y * U_c, \quad c\in\{0,1\},
\end{equation}
where
\begin{equation}
\label{eq:sobel-kernels}
K_x=
\begin{bmatrix}
-1 & 0 & 1\\
-2 & 0 & 2\\
-1 & 0 & 1
\end{bmatrix},
\quad
K_y=
\begin{bmatrix}
-1 & -2 & -1\\
0 & 0 & 0\\
1 & 2 & 1
\end{bmatrix}.
\end{equation}
The structure tensor components are
\begin{equation}
\label{eq:structure-tensor-components}
\begin{gathered}
J_{xx}(\mathbf{U}) = \sum_{c=0}^{1} U_{x,c}^2, \ 
J_{yy}(\mathbf{U}) = \sum_{c=0}^{1} U_{y,c}^2,\\
J_{xy}(\mathbf{U}) = \sum_{c=0}^{1} U_{x,c}U_{y,c}.
\end{gathered}
\end{equation}
We then use a double-angle representation to obtain a normalized orientation descriptor:
\begin{equation}
\label{eq:local-orientation-descriptor}
\mathbf{O}(\mathbf{U})
=
\frac{
\left(
J_{xx}(\mathbf{U})-J_{yy}(\mathbf{U}),
2J_{xy}(\mathbf{U})
\right)
}{
\sqrt{
\left(J_{xx}(\mathbf{U})-J_{yy}(\mathbf{U})\right)^2
+
\left(2J_{xy}(\mathbf{U})\right)^2
+\varepsilon
}
}.
\end{equation}
where $\varepsilon$ is a small positive constant, preventing division by zero. The corresponding orientation energy is
\begin{equation}
\label{eq:orientation-energy}
\mathbf{E}(\mathbf{U})
=
\sqrt{
\left(J_{xx}(\mathbf{U})-J_{yy}(\mathbf{U})\right)^2
+
\left(2J_{xy}(\mathbf{U})\right)^2
+\varepsilon
}.
\end{equation}

\subsubsection{Direction-aware boundary regularization}

Prediction and target orientation descriptors are computed as
\begin{equation}
\label{eq:prediction-target-orientation}
\mathbf{O}_p=\mathbf{O}(\mathbf{P}), \quad \mathbf{O}_y=\mathbf{O}(\mathbf{Y}^{\mathrm{sm}}).
\end{equation}
The normalized target orientation energy is
\begin{equation}
\label{eq:normalized-target-orientation-energy}
\hat{\mathbf{E}}_y
=
\frac{
\mathbf{E}(\mathbf{Y}^{\mathrm{sm}})
}{
\max \mathbf{E}(\mathbf{Y}^{\mathrm{sm}})+\varepsilon
},
\end{equation}
which is combined with the boundary-support mask:
\begin{equation}
\label{eq:direction-weight-map}
\mathbf{W}_{\mathrm{dir}} = \mathbf{B}\odot \hat{\mathbf{E}}_y.
\end{equation}
The direction-aware loss is
\begin{equation}
\label{eq:direction-aware-loss}
\mathcal{L}_{\mathrm{dir}}
=
\frac{
\sum_{u,v}
\mathbf{W}_{\mathrm{dir}}(u,v)
\left[
1-
\left\langle
\mathbf{O}_p(u,v),\mathbf{O}_y(u,v)
\right\rangle
\right]
}{
\sum_{u,v}\mathbf{W}_{\mathrm{dir}}(u,v)+\varepsilon
},
\end{equation}
where $\langle\cdot,\cdot\rangle$ is inner product between normalized orientation vectors. Better orientation alignment yields a smaller loss. 

\subsection{Saddle-aware Suppression}
\label{sec:resolution-dependent-saddle}

While the direction-aware loss improves boundary orientation consistency, LR imagery still struggles to separate adjacent buildings. Therefore, we introduce a \textbf{saddle-aware suppression loss} to penalize false positives in building gaps.

\subsubsection{Saddle mask construction}

Let $\mathbf{Y}\in\{0,1\}^{H\times W}$ denote the binary label map. Foreground and background masks are
\begin{equation}
\label{eq:saddle-fg-bg-mask}
\mathbf{Y}_{\mathrm{fg}}=\mathbb{I}(\mathbf{Y}=1), \quad \mathbf{Y}_{\mathrm{bg}}=\mathbb{I}(\mathbf{Y}=0).
\end{equation}
A saddle pixel is defined as a background pixel supported by building pixels from two opposite directions. We detect such pixels using eight fixed $5\times5$ directional kernels for horizontal, vertical, and diagonal directions. For direction $d$, the foreground support response is
\begin{equation}
\label{eq:saddle-direction-response}
R_d(u,v)
=
\sum_{(\Delta u,\Delta v)\in\Omega_d}
\mathbf{Y}_{\mathrm{fg}}(u+\Delta u,v+\Delta v),
\end{equation}
where $\Omega_d$ contains two offsets along direction $d$. For example,
\begin{equation}
\label{eq:saddle-horizontal-support}
\Omega_{\mathrm{l}}=\{(0,-2),(0,-1)\}, \quad \Omega_{\mathrm{r}}=\{(0,1),(0,2)\}.
\end{equation}
The support indicator is
\begin{equation}
\label{eq:saddle-direction-hit}
H_d(u,v)=\mathbb{I}\big(R_d(u,v)>\tau_{\mathrm{sad}}\big),
\end{equation}
where $\tau_{\mathrm{sad}}=0.5$. Opposite-side support is then checked by
\begin{equation}
\label{eq:saddle-opposite-conditions}
\begin{gathered}
C_{\mathrm{lr}} = H_{\mathrm{l}}\wedge H_{\mathrm{r}},\ 
C_{\mathrm{ud}} = H_{\mathrm{u}}\wedge H_{\mathrm{d}},\\
C_{\mathrm{diag1}} = H_{\mathrm{ul}}\wedge H_{\mathrm{dr}},\
C_{\mathrm{diag2}} = H_{\mathrm{ur}}\wedge H_{\mathrm{dl}}.
\end{gathered}
\end{equation}
The final saddle mask is
\begin{equation}
\label{eq:saddle-mask}
\mathbf{M}_{\mathrm{sad}}(u,v)
=
\mathbf{Y}_{\mathrm{bg}}(u,v)
\cdot
\mathbb{I}
\left(
C_{\mathrm{lr}}\vee C_{\mathrm{ud}}
\vee C_{\mathrm{diag1}}\vee C_{\mathrm{diag2}}
\right).
\end{equation}
Thus, $\mathbf{M}_{\mathrm{sad}}$ selects background pixels between nearby buildings and is used only as a non-gradient spatial weighting mask.

\subsubsection{Saddle-weighted Dice loss}

Let $\mathbf{P}_{\mathrm{fg}}\in[0,1]^{H\times W}$ denote the predicted building probability. For resolution group $r\in\{\mathrm{HR},\mathrm{LR}\}$, the false-positive weight is
\begin{equation}
\label{eq:saddle-fp-weight}
\mathbf{W}_{\mathrm{fp}}^{(r)} = 1+\alpha_r\mathbf{M}_{\mathrm{sad}},
\end{equation}
where $\alpha_{\mathrm{HR}}=0$ and $\alpha_{\mathrm{LR}}>0$. The Dice terms are
\begin{equation}
\label{eq:saddle-dice-tp-fp-fn}
\begin{gathered}
TP = \sum \mathbf{P}_{\mathrm{fg}}\mathbf{Y}_{\mathrm{fg}},\\
FP^{(r)} = \sum \mathbf{P}_{\mathrm{fg}}\mathbf{Y}_{\mathrm{bg}}\mathbf{W}_{\mathrm{fp}}^{(r)},\\
FN = \sum (1-\mathbf{P}_{\mathrm{fg}})\mathbf{Y}_{\mathrm{fg}}.
\end{gathered}
\end{equation}
Only the false-positive term is reweighted, because the goal is to suppress false building activations in background gaps. The saddle-aware loss is
\begin{equation}
\label{eq:saddle-aware-loss}
\mathcal{L}_{\mathrm{sad}}^{(r)}
=
1-
\frac{2TP+\varepsilon}
{2TP+FP^{(r)}+FN+\varepsilon}.
\end{equation}
When $\alpha_{\mathrm{HR}}=0$, this reduces to standard soft Dice for HR samples; for LR samples, $\alpha_{\mathrm{LR}}>0$ explicitly penalizes false activations in narrow inter-building gaps.

\subsection{Overall Objective}

The final training objective is
\begin{equation}
\label{eq:overall-objective}
\mathcal{L}^{(r)}
=
\mathcal{L}_{\mathrm{ce}}
+
\lambda_{\mathrm{dir}}\mathcal{L}_{\mathrm{dir}}
+
\mathcal{L}_{\mathrm{sad}}^{(r)},
\quad
r\in\{\mathrm{HR},\mathrm{LR}\}.
\end{equation}
Here two hyper-parameters are required, $\lambda_{\mathrm{dir}}$ for direction-aware boundary regularization and $\alpha_{\mathrm{LR}}$ for LR saddle-aware suppression, with $\alpha_{\mathrm{HR}}=0$ by default. Overall, direction-aware loss regularizes boundary orientation consistency across all resolutions, while saddle-aware loss strengthens inter-building gap preservation under LR conditions.

\section{Experiments}
\label{sec:experiments}

\begin{table*}[h]
\centering
\caption{Training datasets in our experiments. All datasets are converted to binary building extraction labels.}
\label{tab:datasets}
\resizebox{0.98\textwidth}{!}{
\begin{tabular}{llcclccccc}
\toprule
Dataset & Region & Type & Sensor/source & GSD & Patch size & Train & Val & Test & Split ratio \\
\midrule
Potsdam & Germany & Aerial & Orthophoto RGB & $0.05\,\mathrm{m}$ & $512 \times 512$ & 4,377 & 547 & 548 & $8:1:1$ \\
INRIA & Europe/USA & Aerial & Orthophoto RGB & $0.30\,\mathrm{m}$ & $512 \times 512$ & 14,400 & 1,800 & 1,800 & $8:1:1$ \\
LoveDA & China & Mixed & Google Earth & $0.30\,\mathrm{m}$ & $512 \times 512$ & 10,088 & 3,338 & 3,338 & $3:1:1$ \\
SpaceNet2 & Multi-city & Satellite & WorldView-3 & $0.30\,\mathrm{m}$ & $512 \times 512$ & 8,474 & 1,059 & 1,060 & $8:1:1$ \\
LandCover.ai & Poland & Aerial & Orthophoto RGB & $0.30\,\mathrm{m}$ & $512 \times 512$ & 8,539 & 1,067 & 1,068 & $8:1:1$ \\
OEM & Global & Mixed & Aerial/satellite/UAV, multi-source & $0.25$-$0.5\,\mathrm{m}$ & $512 \times 512$ & 7,461 & 932 & 934 & $8:1:1$ \\
ORBITaLNet & Global & Satellite & Maxar VHR, mainly WorldView-2/3 & $0.47\,\mathrm{m}$ & $512 \times 512$ & 115,443 & 6,413 & 6,414 & $18:1:1$ \\
Alabama & USA & Satellite & Bing Maps & $0.50\,\mathrm{m}$ & $512 \times 512$ & 32,640 & 4,080 & 4,080 & $8:1:1$ \\
WHU-Mix & New Zealand & Aerial & LINZ aerial imagery & $0.50\,\mathrm{m}$ & $512 \times 512$ & 39,346 & 4,381 & 8,402 & $9:1:2$ \\
GF-7 & China & Satellite & GaoFen-7 & $0.65\,\mathrm{m}$ & $512 \times 512$ & 3,106 & 1,034 & 1,035 & $3:1:1$ \\
Planet & Global & Satellite & PlanetScope & $4.80\,\mathrm{m}$ & $512 \times 512$ & 80,000 & 10,000 & 10,000 & $8:1:1$ \\
Sentinel-2 & Global & Satellite & Sentinel-2 RGB & $10\,\mathrm{m}$ & $512 \times 512$ & 40,000 & 5,000 & 5,000 & $8:1:1$ \\
\bottomrule
\end{tabular}
}
\vspace{-0.2cm}
\end{table*}

\subsection{Training Datasets and Experimental Settings}
\label{sec:datasets_settings}

\paragraph{Training Datasets} To train UniBuild across different spatial resolutions, sensors, and geographic domains, we collect public and private building extraction datasets and unify their annotations into a binary label space of building and background, where unlabeled or invalid pixels are ignored.
The training datasets cover very-high-resolution aerial imagery, high-resolution satellite imagery, and medium-resolution satellite imagery. HR datasets include Potsdam \citep{rottensteiner2012isprs}, INRIA \citep{maggiori2017inria}, Alabama \citep{cao2022alabama}, OpenEarthMap \citep{xia2023openearthmap}, LoveDA \citep{wang2021loveda}, GF-7 \citep{chen2024gf7}, WHU-Mix \citep{luo2023whumix}, SpaceNet2 \citep{vanetten2018spacenet}, LandCover.ai \citep{boguszewski2021landcoverai}, and ORBITaL-Net \citep{swan2025orbitalnet}. In addition, we construct two large-scale building extraction datasets from Planet imagery \citep{planetTeam2017,zhu2025globalbuildingatlas} and Sentinel-2 (ST-2) imagery \citep{drusch2012sentinel} over globally distributed urban areas, with building annotations derived from OpenStreetMap (OSM). Although OSM-derived labels contain inevitable noise, including missing buildings, outdated annotations, and image--label misalignment, they provide valuable large-scale supervision for geographically diverse LR scenarios. Overall, the training data span 0.05\,m to 10\,m GSD, enabling UniBuild to learn building representations across substantially different object scales and imaging conditions, which are summarized in Table \ref{tab:datasets}.

\paragraph{Experimental Settings} For a fair comparison, all competing models are trained on the same single-dataset splits and evaluated using the same validation protocol, image size, data augmentation, optimizer settings, and metrics. All the models use a $512 \times 512$ input crop size. Each model is trained for 50 epochs using AdamW, with a base learning rate of $5 \times 10^{-6}$, a decoder learning-rate multiplier of 10, and a weight decay of 0.01. The batch size is set to 10. For multi-dataset training, datasets are grouped according to their spatial resolutions. For both HR and LR images, we first apply weak geometric augmentations, including random scaling, cropping, and flipping, followed by stronger intensity augmentations, such as color jittering, Gaussian blurring, sharpening, and grayscale conversion.
We evaluate semantic building extraction using building IoU and F1 score. Since boundary quality is particularly important for building footprint extraction, we further report Boundary-IoU and Boundary-F1. The best checkpoint is selected based on the average of IoU and Boundary-IoU, jointly considering region accuracy and boundary quality.

\subsection{Comparison With SOTA Building Extraction Models}

\begin{table}[h]
\centering
\caption{Comparison with some representative building extraction models and UniBuild under single-dataset training.}
\label{tab:sota-comparison}
\scriptsize
\setlength{\tabcolsep}{4.5pt}
\renewcommand{\arraystretch}{0.95}
\resizebox{0.484\textwidth}{!}{
\begin{tabular}{llcc|cc|cc}
\toprule
Dataset & Method
& IoU & F1
& B-IoU & B-F1
& \cellcolor{red!8}M$_{\mathrm{IoU}}$
& \cellcolor{red!8}M$_{\mathrm{F1}}$ \\
\midrule
\multirow{8}{*}{INRIA}
& U-Net        & 75.31 & 85.92 & 51.59 & 55.04 & \cellcolor{red!6}63.45 & \cellcolor{red!6}70.48 \\
& DeepLabV3+   & 75.92 & 86.31 & 50.71 & 53.88 & \cellcolor{red!6}63.32 & \cellcolor{red!6}70.10 \\
& BuildFormer  & 77.59 & 87.38 & 55.34 & 58.52 & \cellcolor{red!6}66.46 & \cellcolor{red!6}72.95 \\
& SegFormer    & 77.52 & 87.34 & 53.03 & 56.16 & \cellcolor{red!6}65.27 & \cellcolor{red!6}71.75 \\
& BIENet       & 76.93 & 86.96 & 55.11 & 58.46 & \cellcolor{red!6}66.02 & \cellcolor{red!6}72.71 \\
& BOMSC-Net    & 76.31 & 86.57 & 53.18 & 56.45 & \cellcolor{red!6}64.75 & \cellcolor{red!6}71.51 \\
& DINOv3-B-DPT & 82.17 & 90.21 & 66.91 & 70.04 & \cellcolor{red!6}74.54 & \cellcolor{red!6}80.12 \\
& UniBuild     & \textbf{83.29} & \textbf{90.88} & \textbf{70.86} & \textbf{74.09}
& \cellcolor{red!7}\textbf{77.08} & \cellcolor{red!7}\textbf{82.49} \\
\midrule
\multirow{8}{*}{GF-7}
& U-Net        & 63.14 & 77.41 & 53.01 & 56.78 & \cellcolor{red!6}58.08 & \cellcolor{red!6}67.09 \\
& DeepLabV3+   & 63.41 & 77.61 & 50.70 & 53.70 & \cellcolor{red!6}57.05 & \cellcolor{red!6}65.66 \\
& BuildFormer  & 69.04 & 81.68 & 60.12 & 63.16 & \cellcolor{red!6}64.58 & \cellcolor{red!6}72.42 \\
& SegFormer    & 68.33 & 81.19 & 56.48 & 59.55 & \cellcolor{red!6}62.40 & \cellcolor{red!6}70.37 \\
& BIENet       & 66.98 & 80.23 & 58.71 & 62.03 & \cellcolor{red!6}62.85 & \cellcolor{red!6}71.13 \\
& BOMSC-Net    & 65.81 & 79.38 & 56.04 & 59.28 & \cellcolor{red!6}60.93 & \cellcolor{red!6}69.33 \\
& DINOv3-B-DPT & 76.27 & 86.54 & 71.10 & 74.01 & \cellcolor{red!6}73.69 & \cellcolor{red!6}80.28 \\
& UniBuild     & \textbf{78.68} & \textbf{88.07} & \textbf{75.76} & \textbf{79.04}
& \cellcolor{red!7}\textbf{77.22} & \cellcolor{red!7}\textbf{83.56} \\
\midrule
\multirow{8}{*}{WHU-Mix}
& U-Net        & 78.49 & 87.95 & 55.83 & 59.58 & \cellcolor{red!6}67.16 & \cellcolor{red!6}73.76 \\
& DeepLabV3+   & 79.48 & 88.57 & 56.42 & 59.79 & \cellcolor{red!6}67.95 & \cellcolor{red!6}74.18 \\
& BuildFormer  & 81.03 & 89.52 & 61.43 & 64.83 & \cellcolor{red!6}71.23 & \cellcolor{red!6}77.18 \\
& SegFormer    & 80.70 & 89.32 & 59.44 & 62.84 & \cellcolor{red!6}70.07 & \cellcolor{red!6}76.08 \\
& BIENet       & 79.93 & 88.85 & 60.99 & 64.62 & \cellcolor{red!6}70.46 & \cellcolor{red!6}76.73 \\
& BOMSC-Net    & 79.21 & 88.40 & 58.22 & 61.87 & \cellcolor{red!6}68.71 & \cellcolor{red!6}75.13 \\
& DINOv3-B-DPT & 84.64 & 91.68 & 70.17 & 73.51 & \cellcolor{red!6}77.41 & \cellcolor{red!6}82.59 \\
& UniBuild     & \textbf{85.26} & \textbf{92.05} & \textbf{70.96} & \textbf{74.42}
& \cellcolor{red!7}\textbf{78.11} & \cellcolor{red!7}\textbf{83.23} \\
\midrule
\multirow{8}{*}{Planet}
& U-Net        & 37.84 & 54.90 & 35.57 & 37.08 & \cellcolor{red!6}36.70 & \cellcolor{red!6}45.99 \\
& DeepLabV3+   & 39.99 & 57.13 & 34.92 & 36.46 & \cellcolor{red!6}37.46 & \cellcolor{red!6}46.80 \\
& BuildFormer  & 40.30 & 57.45 & 35.77 & 37.23 & \cellcolor{red!6}38.04 & \cellcolor{red!6}47.34 \\
& SegFormer    & 42.05 & 59.21 & 36.61 & 38.10 & \cellcolor{red!6}39.33 & \cellcolor{red!6}48.65 \\
& BIENet       & 38.34 & 55.43 & 34.51 & 35.91 & \cellcolor{red!6}36.43 & \cellcolor{red!6}45.67 \\
& BOMSC-Net    & 38.66 & 55.77 & 34.49 & 35.98 & \cellcolor{red!6}36.58 & \cellcolor{red!6}45.87 \\
& DINOv3-B-DPT & 45.35 & 62.40 & 42.02 & 43.61 & \cellcolor{red!6}43.69 & \cellcolor{red!6}53.01 \\
& UniBuild     & \textbf{48.25} & \textbf{65.10} & \textbf{47.35} & \textbf{49.30}
& \cellcolor{red!7}\textbf{47.80} & \cellcolor{red!7}\textbf{57.20} \\
\bottomrule
\end{tabular}
}
\vspace{-0.3cm}
\end{table}

\begin{table}[h]
\centering
\caption{Model complexity comparison under a $512\times512$ RGB input. PM and GP denote the number of trainable parameters and GFLOPs for the entire model, respectively.}
\label{tab:model-complexity}
\scriptsize
\setlength{\tabcolsep}{8pt}
\renewcommand{\arraystretch}{0.95}
\begin{tabular}{lcc}
\toprule
Method & PM (M) & GP (G) \\
\midrule
U-Net             & 24.44 & 31.29  \\
DeepLabV3+        & 26.68 & 36.57  \\
BuildFormer       & 32.90 & 81.72  \\
SegFormer         & 27.35 & 56.58  \\
BIENet            & 32.58 & 37.89  \\
BOMSC-Net         & 26.75 & 32.85  \\
DINOv3-B-DPT      & 96.62 & 120.78 \\
UniBuild / DINOv3-B-HR-DPT   & 97.68 & 160.57 \\
\bottomrule
\end{tabular}
\vspace{-0.3cm}
\end{table}

To assess the effectiveness of UniBuild, we compare it with representative general semantic segmentation models and recent building extraction models, using the hyperparameters specified in Sec.~\ref{sec:ablation}. The general segmentation baselines include U-Net \citep{ronneberger2015unet} with a ResNet-34 backbone \citep{he2016deep}, DeepLabV3+ \citep{chen2018encoder} with a ResNet-50 backbone \citep{he2016deep}, and SegFormer \citep{xie2021segformer} with a MiT-B2 backbone. The building-specific baselines include BuildFormer \citep{wang2022buildformer} with a Swin-T backbone \citep{liu2021swin}, BIENet \citep{bienet2024} with a ResNet-50 backbone, and BOMSC-Net \citep{li2022bomscnet} with its original backbone design. We further include DINOv3-B-DPT as a strong foundation-model baseline, which uses the same DINOv3-Base backbone \citep{simeoni2025dinov3} as UniBuild but adopts a standard DPT decoder and CE loss. UniBuild is evaluated as a complete framework, including the proposed HR-DPT decoder and geometry-aware regularization losses. All comparison models are trained with CE loss under their standard segmentation settings. The UniBuild results in Table~\ref{tab:sota-comparison} are obtained from single-dataset checkpoints for each dataset, rather than from the multi-dataset joint training.

\begin{figure*}[h]
    \centering
    \includegraphics[width=1.0\linewidth]{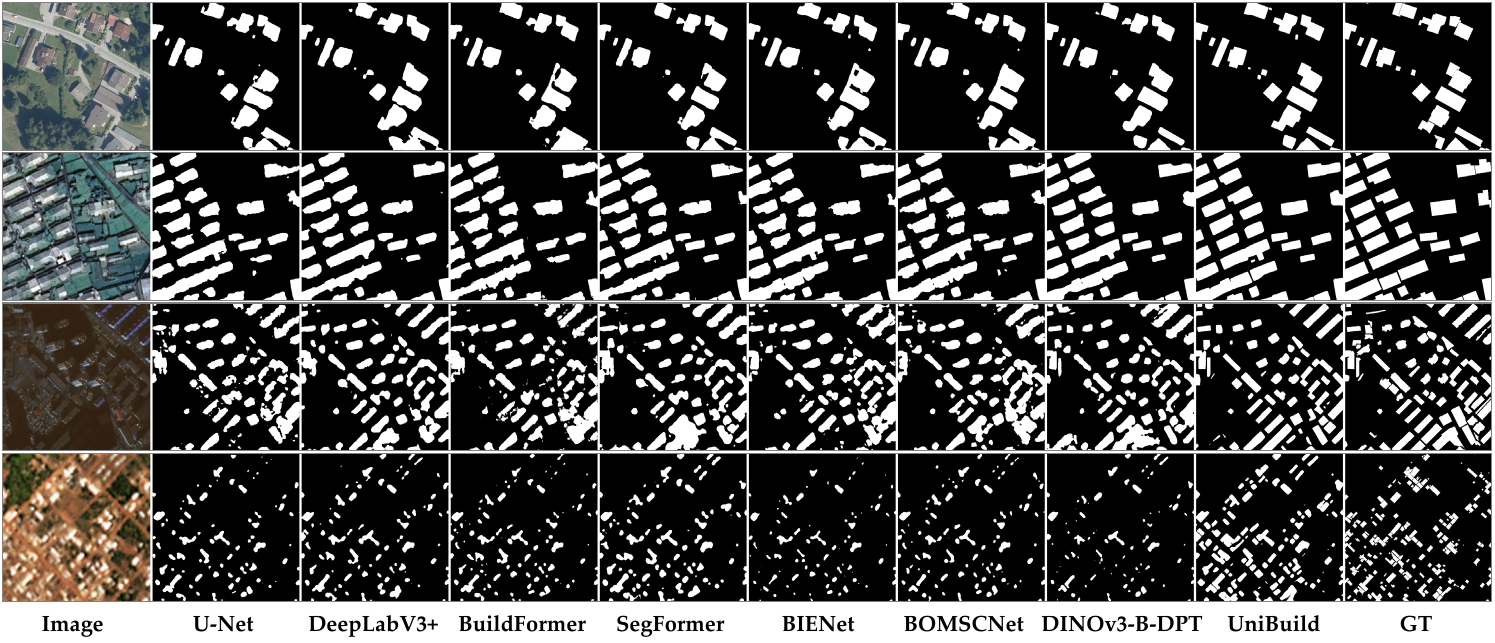}
    \vspace{-0.3cm}
    \caption{Qualitative comparison with competing models, DINOv3-B-DPT, and UniBuild. UniBuild uses the corresponding single-dataset trained checkpoint for each dataset. Rows 1--4 show examples from INRIA, WHU-Mix, GF-7, and Planet (bilinearly upsampled to 1 m), respectively. For each row, the columns present the RGB image, predictions from different methods, and the ground-truth building label.}
    \label{fig:sota-comparison-samples}
    \vspace{-0.1cm}
\end{figure*}

As shown in Table~\ref{tab:sota-comparison}, DINOv3-B-DPT already substantially outperforms most conventional segmentation models and building-specific architectures, demonstrating the strong representation ability of the pretrained visual foundation backbone. Built upon the same DINOv3-Base backbone, UniBuild further achieves the best overall performance across the four representative datasets, with clear gains in both region-level and boundary-aware metrics. The improvements in B-IoU and B-F1 are particularly important, as accurate boundaries are essential for downstream tasks such as building polygonization. These results indicate that the performance gains of UniBuild come not only from the foundation backbone, but also from the proposed HR-DPT decoder and geometry-aware regularization. The isolated effects of these two components are further analyzed in Sec.~\ref{sec:ablation}.
The qualitative examples in Fig.~\ref{fig:sota-comparison-samples} further support these findings, showing that UniBuild produces more complete masks, sharper boundaries, and clearer separation between adjacent buildings, especially for small buildings, dense layouts, and LR imagery.

\subsection{Single-dataset vs. Multi-dataset Training}
\label{sec:multidataset-training}

\begin{table}[h]
\centering
\caption{Comparison between single-dataset DINOv3-B-DPT models and UniBuild. ``B-IoU'' and ``B-F1'' denote Boundary-IoU and Boundary-F1, respectively. ``M$_{\mathrm{IoU}}$'' is the mean of IoU and B-IoU, while ``M$_{\mathrm{F1}}$'' is the mean of F1 and B-F1.}
\label{tab:dpt-unibuild}
\scriptsize
\setlength{\tabcolsep}{5.0pt}
\renewcommand{\arraystretch}{0.95}
\resizebox{0.484\textwidth}{!}{
\begin{tabular}{llcc|cc|cc}
\toprule
Dataset & Method
& IoU & F1
& B-IoU & B-F1
& \cellcolor{red!10}M$_{\mathrm{IoU}}$
& \cellcolor{red!10}M$_{\mathrm{F1}}$ \\
\midrule
\multirow{2}{*}{Potsdam}
& DINOv3-B-DPT & 90.63 & 95.11 & 51.56 & 54.85
& \cellcolor{red!7}71.10 & \cellcolor{red!7}74.98 \\
& UniBuild & \textbf{93.04} & \textbf{96.39}
& \textbf{60.79} & \textbf{64.80}
& \cellcolor{red!7}\textbf{76.92}
& \cellcolor{red!7}\textbf{80.59} \\
\midrule
\multirow{2}{*}{INRIA}
& DINOv3-B-DPT & 82.17 & 90.21 & 66.91 & 70.04
& \cellcolor{red!7}74.54 & \cellcolor{red!7}80.12 \\
& UniBuild & \textbf{83.34} & \textbf{90.92}
& \textbf{69.30} & \textbf{72.71}
& \cellcolor{red!7}\textbf{76.32}
& \cellcolor{red!7}\textbf{81.81} \\
\midrule
\multirow{2}{*}{Alabama}
& DINOv3-B-DPT & 79.01 & 88.27 & 80.42 & 83.09
& \cellcolor{red!7}79.72 & \cellcolor{red!7}85.68 \\
& UniBuild & \textbf{80.51} & \textbf{89.20}
& \textbf{82.94} & \textbf{85.54}
& \cellcolor{red!7}\textbf{81.72}
& \cellcolor{red!7}\textbf{87.37} \\
\midrule
\multirow{2}{*}{OEM}
& DINOv3-B-DPT & 83.03 & 90.73 & 74.83 & 77.95
& \cellcolor{red!7}78.93 & \cellcolor{red!7}84.34 \\
& UniBuild & \textbf{84.22} & \textbf{91.43}
& \textbf{78.07} & \textbf{81.54}
& \cellcolor{red!7}\textbf{81.14}
& \cellcolor{red!7}\textbf{86.49} \\
\midrule
\multirow{2}{*}{WHU-Mix}
& DINOv3-B-DPT & 84.64 & 91.68 & 70.17 & 73.51
& \cellcolor{red!7}77.41 & \cellcolor{red!7}82.59 \\
& UniBuild & \textbf{85.42} & \textbf{92.13}
& \textbf{71.18} & \textbf{74.58}
& \cellcolor{red!7}\textbf{78.30}
& \cellcolor{red!7}\textbf{83.35} \\
\midrule
\multirow{2}{*}{GF-7}
& DINOv3-B-DPT & 76.27 & 86.54 & 71.10 & 74.01
& \cellcolor{red!7}73.69 & \cellcolor{red!7}80.28 \\
& UniBuild & \textbf{80.23} & \textbf{89.03}
& \textbf{78.03} & \textbf{81.32}
& \cellcolor{red!7}\textbf{79.13}
& \cellcolor{red!7}\textbf{85.17} \\
\midrule
\multirow{2}{*}{LoveDA}
& DINOv3-B-DPT & 71.00 & 83.04 & 30.29 & 32.69
& \cellcolor{red!7}50.64 & \cellcolor{red!7}57.87 \\
& UniBuild & \textbf{71.64} & \textbf{83.48}
& \textbf{35.92} & \textbf{38.64}
& \cellcolor{red!7}\textbf{53.78}
& \cellcolor{red!7}\textbf{61.06} \\
\midrule
\multirow{2}{*}{SpaceNet2}
& DINOv3-B-DPT & 79.69 & 88.70 & 66.19 & 69.24
& \cellcolor{red!7}72.94 & \cellcolor{red!7}78.97 \\
& UniBuild & \textbf{81.54} & \textbf{89.83}
& \textbf{71.38} & \textbf{74.46}
& \cellcolor{red!7}\textbf{76.46}
& \cellcolor{red!7}\textbf{82.14} \\
\midrule
\multirow{2}{*}{LandCover.ai}
& DINOv3-B-DPT & 84.14 & 91.39 & 77.64 & 80.09
& \cellcolor{red!7}80.89 & \cellcolor{red!7}85.74 \\
& UniBuild & \textbf{87.31} & \textbf{93.23}
& \textbf{86.40} & \textbf{88.51}
& \cellcolor{red!7}\textbf{86.86}
& \cellcolor{red!7}\textbf{90.87} \\
\midrule
\multirow{2}{*}{ORBITaLNet}
& DINOv3-B-DPT & 85.39 & 92.12 & 80.02 & 83.40
& \cellcolor{red!7}82.70 & \cellcolor{red!7}87.76 \\
& UniBuild & \textbf{85.60} & \textbf{92.24}
& \textbf{80.76} & \textbf{84.02}
& \cellcolor{red!7}\textbf{83.18}
& \cellcolor{red!7}\textbf{88.13} \\
\midrule
\multirow{2}{*}{Planet}
& DINOv3-B-DPT & 45.34 & 62.40 & 42.02 & 43.61
& \cellcolor{red!7}43.68 & \cellcolor{red!7}53.00 \\
& UniBuild & \textbf{47.91} & \textbf{64.78}
& \textbf{43.91} & \textbf{45.84}
& \cellcolor{red!7}\textbf{45.91}
& \cellcolor{red!7}\textbf{55.31} \\
\midrule
\multirow{2}{*}{ST-2}
& DINOv3-B-DPT & 27.04 & 43.04 & 21.57 & 22.12
& \cellcolor{red!7}24.31 & \cellcolor{red!7}32.58 \\
& UniBuild & \textbf{34.52} & \textbf{51.32}
& \textbf{28.56} & \textbf{29.73}
& \cellcolor{red!7}\textbf{31.54}
& \cellcolor{red!7}\textbf{40.52} \\
\midrule
\multirow{2}{*}{Average}
& DINOv3-B-DPT & 74.03 & 83.60 & 61.06 & 63.72
& \cellcolor{red!7}67.54 & \cellcolor{red!7}73.66 \\
& UniBuild & \textbf{76.27} & \textbf{85.33}
& \textbf{65.60} & \textbf{68.47}
& \cellcolor{red!7}\textbf{70.94}
& \cellcolor{red!7}\textbf{76.90} \\
\bottomrule
\end{tabular}
}
\end{table}

\begin{figure}[h]
    \centering
    \includegraphics[width=0.48\textwidth]
    {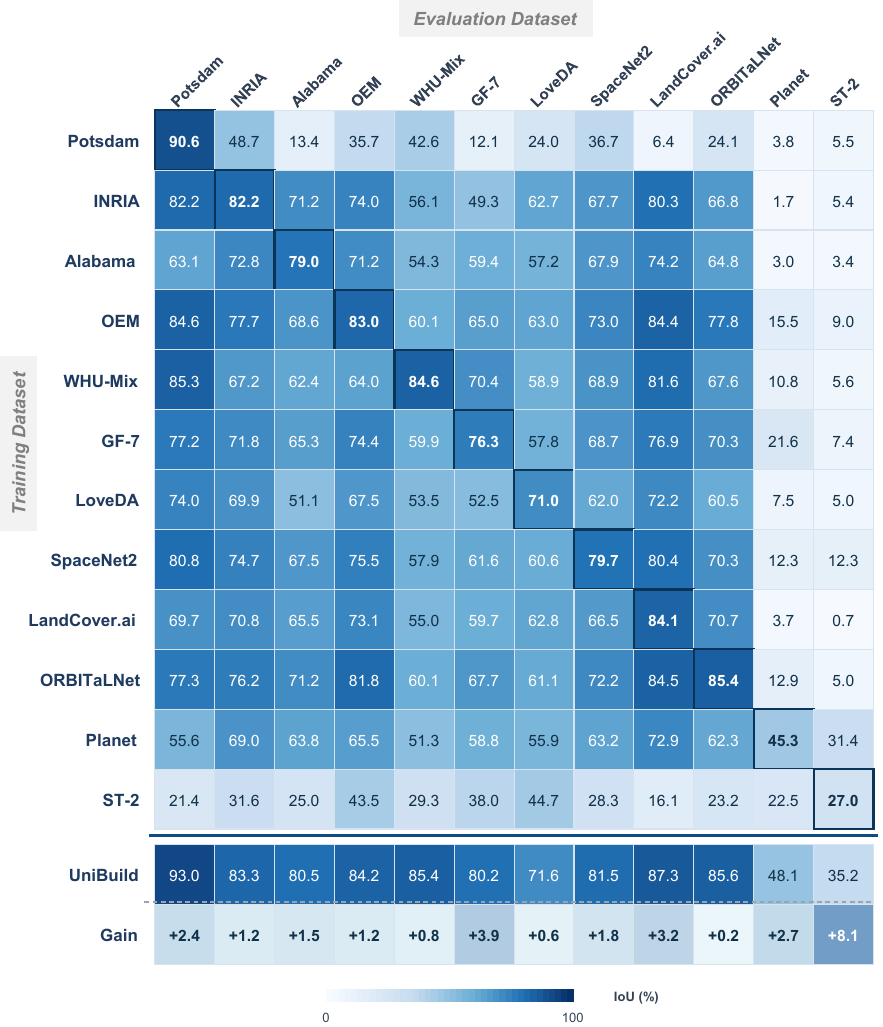}
    \caption{Cross-dataset IoU comparison between single-dataset DINOv3-B-DPT models and multi-dataset UniBuild. Rows indicate training datasets and columns indicate evaluation datasets. Gain reports the IoU improvement of UniBuild over the diagonal DINOv3-B-DPT result.}
    \label{fig:cross-dataset-generalization}
    \vspace{-0.3cm}
\end{figure}

Two settings are compared: (1) single-dataset DINOv3-B-DPT trained independently on each dataset with CE loss; and (2) UniBuild, jointly trained on all 12 datasets with our HR-DPT decoder and geometry-aware regularizations. As shown in Table~\ref{tab:dpt-unibuild}, UniBuild improves the average IoU and B-IoU across the 12 datasets from 74.03 and 61.06 to 76.27 and 65.60, respectively,  demonstrating that UniBuild consistently enhances both region-level accuracy and boundary quality compared with independently trained DINOv3-B-DPT models.

We further evaluate the cross-dataset transferability of single-dataset DINOv3-B-DPT models and compare them with the multi-dataset UniBuild in Fig.~\ref{fig:cross-dataset-generalization}. UniBuild achieves positive gains across all datasets, indicating that multi-dataset joint training does not compromise dataset-specific performance but enables one unified model to maintain consistently strong results across multi-source RS images. These improvements can be attributed to both the proposed architecture and the diversity of multi-dataset training. The HR-DPT decoder and geometry-aware losses enhance detail recovery and boundary regularity, while multi-source datasets introduce broader scale, scene, and annotation variations. Such diversity reduces reliance on dataset-specific object sizes and background co-occurrence patterns, improves robustness to boundary ambiguity and label inconsistency, and helps the foundation backbone preserve more transferable building representations.

\subsection{Ablation Study}
\label{sec:ablation}

\subsubsection{Overall Module Contribution}

We first provide an overall ablation study on two HR datasets, INRIA and GF-7, and two MR/LR datasets, Planet and ST-2. All variants use the same DINOv3-Base backbone and follow identical data splits and training protocol. DPT serves as the baseline decoder, HR-DPT denotes the proposed decoder without geometry-aware regularization, and HR-DPT + Dir + Sad denotes the full UniBuild configuration. For direction-aware regularization, we set $\lambda_{\mathrm{dir}}=0.5$. For saddle-aware regularization, we adopt the resolution-adaptive strategy in Sec.~\ref{sec:resolution-dependent-saddle}, where the saddle penalty is disabled for HR batches and set to $\alpha_{\mathrm{LR}}=15$ for LR batches. As summarized in Table~\ref{tab:ablation-study}, HR-DPT improves both region and boundary metrics over DPT, confirming the benefit of HR detail recovery from spatially compressed VFM features. Direction-aware regularization further improves boundary regularity, while saddle-aware suppression brings clear gains on LR datasets by reducing false-positive connections between adjacent buildings. In particular, M$_{\mathrm{IoU}}$ increases from 45.77 to 47.80 on Planet and from 26.97 to 33.60 on ST-2 after adding saddle-aware suppression. Detailed analyses of each component are provided in the following subsections.

\begin{table}[h]
    \centering
    \caption{Overall ablation study of single-dataset UniBuild. }
    \vspace{-0.1cm}
    \label{tab:ablation-study}
    \scriptsize
    \setlength{\tabcolsep}{4.5pt}
    \renewcommand{\arraystretch}{0.95}
    \resizebox{0.484\textwidth}{!}{
        \begin{tabular}{llcc|cc|cc}
        \toprule
        Dataset & Method
        & IoU & F1
        & B-IoU & B-F1
        & \cellcolor{red!10}M$_{\mathrm{IoU}}$
        & \cellcolor{red!10}M$_{\mathrm{F1}}$ \\
        \midrule
        \multirow{4}{*}{INRIA}
        & DPT & 82.17 & 90.21 & 66.91 & 70.04
        & \cellcolor{red!7}74.54 & \cellcolor{red!7}80.13 \\
        & HR-DPT & 82.47 & 90.39 & 67.81 & 71.15
        & \cellcolor{red!7}75.14 & \cellcolor{red!7}80.77 \\
        & HR-DPT + Dir & 83.13 & 90.79 & 70.54 & 73.79
        & \cellcolor{red!7}76.84 & \cellcolor{red!7}82.29 \\
        & HR-DPT + Dir + Sad
        & \textbf{83.29} & \textbf{90.88}
        & \textbf{70.86} & \textbf{74.09}
        & \cellcolor{red!7}\textbf{77.08}
        & \cellcolor{red!7}\textbf{82.49} \\
        \midrule
        \multirow{4}{*}{GF-7}
        & DPT & 76.27 & 86.54 & 71.10 & 74.01
        & \cellcolor{red!7}73.69 & \cellcolor{red!7}80.28 \\
        & HR-DPT & 77.94 & 87.60 & 74.57 & 77.89
        & \cellcolor{red!7}76.25 & \cellcolor{red!7}82.75 \\
        & HR-DPT + Dir & 78.48 & 87.94 & 75.70 & 78.99
        & \cellcolor{red!7}77.09 & \cellcolor{red!7}83.47 \\
        & HR-DPT + Dir + Sad
        & \textbf{78.68} & \textbf{88.07}
        & \textbf{75.76} & \textbf{79.04}
        & \cellcolor{red!7}\textbf{77.22}
        & \cellcolor{red!7}\textbf{83.56} \\
        \midrule
        \multirow{4}{*}{Planet}
        & DPT & 45.35 & 62.40 & 42.02 & 43.61
        & \cellcolor{red!7}43.69 & \cellcolor{red!7}53.01 \\
        & HR-DPT & 45.71 & 62.74 & 42.64 & 44.25
        & \cellcolor{red!7}44.18 & \cellcolor{red!7}53.50 \\
        & HR-DPT + Dir & 45.44 & 62.48 & 46.10 & 48.02
        & \cellcolor{red!6}45.77 & \cellcolor{red!6}55.25 \\
        & HR-DPT + Dir + Sad
        & \textbf{48.25} & \textbf{65.10}
        & \textbf{47.35} & \textbf{49.30}
        & \cellcolor{red!7}\textbf{47.80}
        & \cellcolor{red!7}\textbf{57.20} \\
        \midrule
        \multirow{4}{*}{ST-2}
        & DPT & 27.05 & 42.58 & 22.53 & 23.07
        & \cellcolor{red!7}24.79 & \cellcolor{red!7}32.83 \\
        & HR-DPT & 28.21 & 44.00 & 23.18 & 23.83
        & \cellcolor{red!7}25.70 & \cellcolor{red!7}33.92 \\
        & HR-DPT + Dir & 26.17 & 41.48 & 27.76 & 29.61
        & \cellcolor{red!7}26.97 & \cellcolor{red!7}35.55 \\
        & HR-DPT + Dir + Sad
        & \textbf{36.73} & \textbf{53.72}
        & \textbf{30.47} & \textbf{31.69}
        & \cellcolor{red!7}\textbf{33.60}
        & \cellcolor{red!7}\textbf{42.71} \\
        \bottomrule
        \end{tabular}
    }
\end{table}

\subsubsection{Decoder Effectiveness and Complexity}

\begin{figure}[h]
    \centering
    \includegraphics[width=0.92\linewidth]{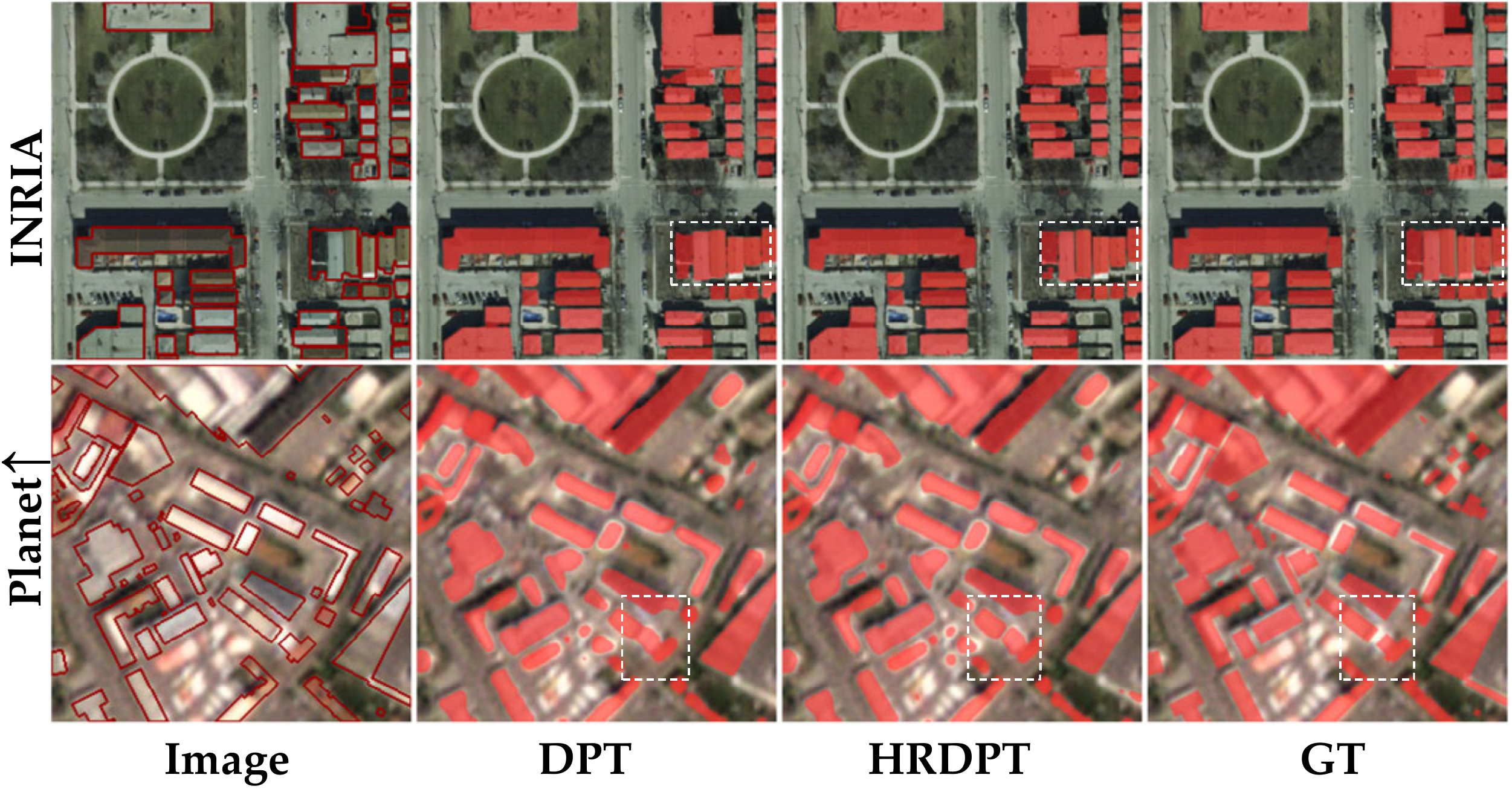}
    \caption{Visual comparison of DPT and HR-DPT trained with CE loss on INRIA and Planet. Planet examples are bilinearly upsampled to 1\,m.}
    \label{fig:ablation_decoder}
\vspace{-0.3cm}
\end{figure}

We compare the decoder-only parameters (PM) and GFLOPs (GP) of UPerNet \citep{xiao2018upernet}, DPT \citep{ranftl2021dpt}, and HR-DPT under the same DINOv3-B backbone with a $512\times512$ input. As shown in Table~\ref{tab:decoder-ablation}, HR-DPT requires 12.0M parameters and 70.1 GFLOPs, slightly higher than DPT but much lower than UPerNet. With this limited additional cost, HR-DPT consistently outperforms DPT and achieves better boundary reconstruction, producing finer building structures and cleaner boundaries under the same CE supervision, as shown in Fig.~\ref{fig:ablation_decoder}.

\begin{table}[h]
    \centering
    \caption{Decoder-only comparison under a $512\times512$ input. PM and GP denote parameter number (M) and FLOPs (G).}
    \label{tab:decoder-ablation}
    \scriptsize
    \setlength{\tabcolsep}{3pt}
    \renewcommand{\arraystretch}{1.1}
    \resizebox{0.49\textwidth}{!}{
    \begin{tabular}{llcc|cc|cc|cc}
    \toprule
    Dataset & Decoder
    & IoU & F1
    & B-IoU & B-F1
    & \cellcolor{red!8}M$_{\mathrm{IoU}}$
    & \cellcolor{red!8}M$_{\mathrm{F1}}$
    & PM (M) & GP (G) \\
    \midrule
    \multirow{3}{*}{INRIA}
    & UPerNet & 82.25 & 90.26 & 66.83 & 69.97
    & \cellcolor{red!6}74.54 & \cellcolor{red!6}80.12 & 38.1 & 213.1 \\
    & DPT & 82.17 & 90.21 & 66.91 & 70.04
    & \cellcolor{red!6}74.54 & \cellcolor{red!6}80.13 & 11.0 & 30.1 \\
    & HR-DPT & \textbf{82.47} & \textbf{90.39}
    & \textbf{67.81} & \textbf{71.15}
    & \cellcolor{red!6}\textbf{75.14}
    & \cellcolor{red!6}\textbf{80.77} & 12.0 & 70.1 \\
    \midrule
    \multirow{3}{*}{GF-7}
    & UPerNet & 76.82 & 86.89 & 72.48 & 75.46
    & \cellcolor{red!6}74.65 & \cellcolor{red!6}81.18 & 38.1 & 213.1 \\
    & DPT & 76.27 & 86.54 & 71.10 & 74.01
    & \cellcolor{red!6}73.69 & \cellcolor{red!6}80.28 & 11.0 & 30.1 \\
    & HR-DPT & \textbf{77.94} & \textbf{87.60}
    & \textbf{74.57} & \textbf{77.89}
    & \cellcolor{red!6}\textbf{76.25}
    & \cellcolor{red!6}\textbf{82.75} & 12.0 & 70.1 \\
    \midrule
    \multirow{3}{*}{Planet}
    & UPerNet & 44.70 & 61.78 & 41.26 & 42.83
    & \cellcolor{red!6}42.98 & \cellcolor{red!6}52.31 & 38.1 & 213.1 \\
    & DPT & 45.35 & 62.40 & 42.02 & 43.61
    & \cellcolor{red!6}43.69 & \cellcolor{red!6}53.01 & 11.0 & 30.1 \\
    & HR-DPT & \textbf{45.71} & \textbf{62.74}
    & \textbf{42.64} & \textbf{44.25}
    & \cellcolor{red!6}\textbf{44.18}
    & \cellcolor{red!6}\textbf{53.50} & 12.0 & 70.1 \\
    \midrule
    \multirow{3}{*}{ST-2}
    & UPerNet & \textbf{28.70} & \textbf{44.60} & 21.37 & 21.91
    & \cellcolor{red!6}25.04 & \cellcolor{red!6}33.26 & 38.1 & 213.1 \\
    & DPT & 27.05 & 42.58 & 22.53 & 23.07
    & \cellcolor{red!6}24.79 & \cellcolor{red!6}32.83 & 11.0 & 30.1 \\
    & HR-DPT & 28.21 & 44.00
    & \textbf{23.18} & \textbf{23.83}
    & \cellcolor{red!6}\textbf{25.70}
    & \cellcolor{red!6}\textbf{33.92} & 12.0 & 70.1 \\
    \bottomrule
    \end{tabular}
    }
    \vspace{-0.1cm}
\end{table}

\subsubsection{Direction-Aware Consistency}

\begin{figure}[h]
    \centering
    \includegraphics[width=0.98\linewidth]{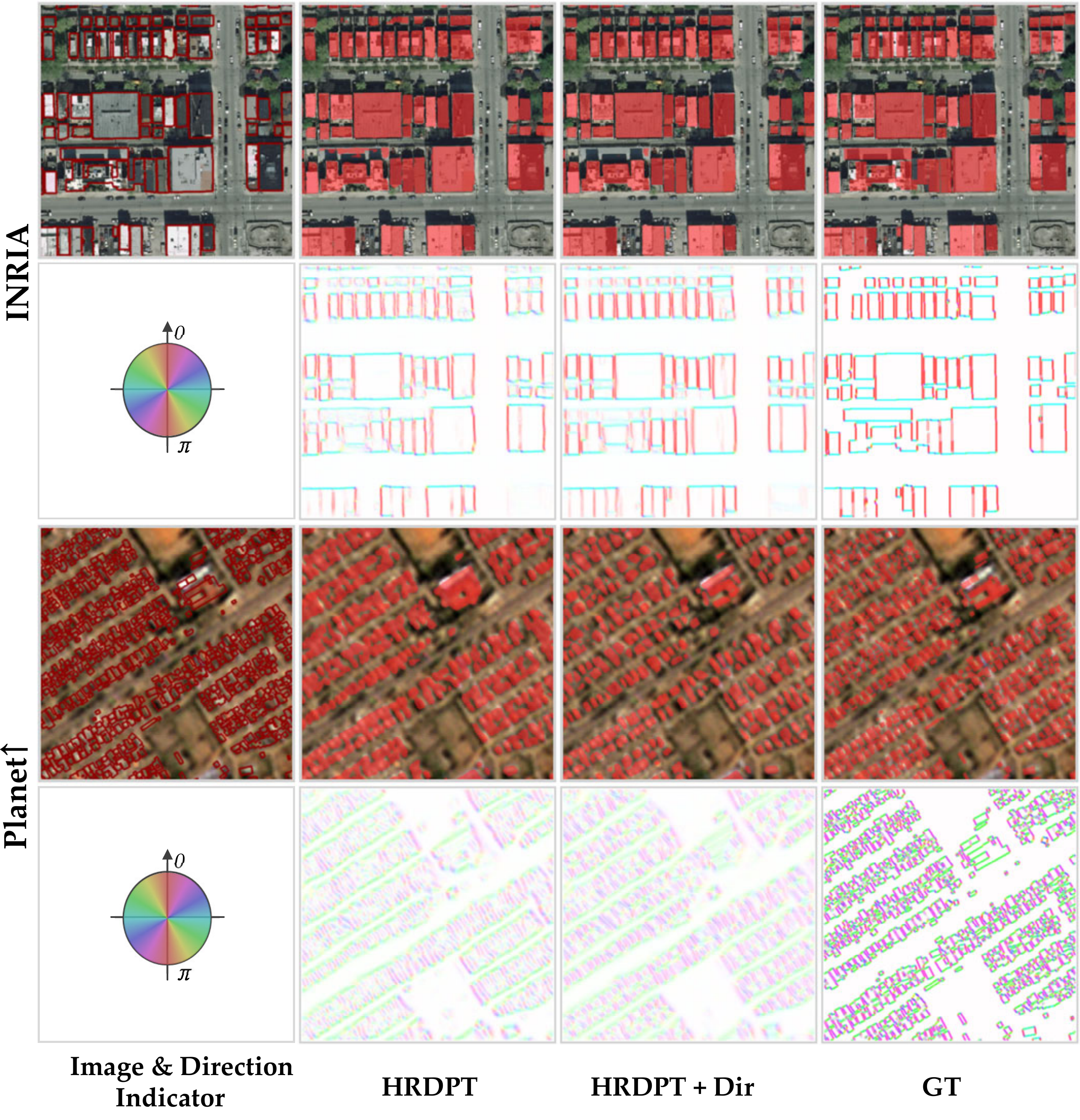}
    \vspace{-0.5cm}
    \caption{Visual comparison of HR-DPT trained with CE loss and CE+Dir on INRIA and Planet. Planet examples are bilinearly upsampled to 1\,m.}
    \label{fig:ablation_direction}
    \vspace{-0.2cm}
\end{figure}

We further study direction-aware regularization on top of HR-DPT. This loss uses structure-tensor-based orientation cues to encourage local boundary consistency and suppress fragmented or zigzag contour responses. As shown in Table~\ref{tab:dir-hyperparam}, it significantly improves boundary-aware metrics. For example, on INRIA, B-IoU and B-F1 increase from 67.81 and 71.15 to 70.87 and 74.09, respectively. This indicates that the orientation constraint effectively improves the geometric consistency of building contours.
On MR/LR datasets, direction-aware regularization also improves boundary metrics, but its effect is more sensitive due to indistinct boundaries and mixed pixels. Therefore, this loss is mainly used as a boundary refinement term. Considering the balance between boundary quality and region completeness, we set $\lambda_{\mathrm{dir}}=0.5$ as the default setting. The results in Fig.~\ref{fig:ablation_direction} further show that direction-aware regularization produces straighter and more coherent building boundaries with fewer irregular contour responses.

\begin{table}[h]
    \centering
    \caption{Effect of direction-aware regularization weight $\lambda_{\mathrm{dir}}$. ``w/o Dir'' denotes without direction-aware regularization.}
    \vspace{-0.2cm}
    \label{tab:dir-hyperparam}
    \scriptsize
    \setlength{\tabcolsep}{6.5pt}
    \renewcommand{\arraystretch}{0.95}
    \resizebox{0.484\textwidth}{!}{
    \begin{tabular}{llcc|cc|cc}
    \toprule
    Dataset & $\lambda_{\mathrm{dir}}$
    & IoU & F1
    & B-IoU & B-F1
    & \cellcolor{red!8}M$_{\mathrm{IoU}}$
    & \cellcolor{red!8}M$_{\mathrm{F1}}$ \\
    \midrule
    \multirow{6}{*}{INRIA}
    & w/o Dir & 82.47 & 90.39 & 67.81 & 71.15
    & \cellcolor{red!6}75.14 & \cellcolor{red!6}80.77 \\
    & 0.2 & 82.93 & 90.67 & 70.15 & 73.42
    & \cellcolor{red!6}76.54 & \cellcolor{red!6}82.05 \\
    & 0.5 & 83.13 & 90.79 & 70.54 & 73.79
    & \cellcolor{red!6}76.84 & \cellcolor{red!6}82.29 \\
    & 1.0 & \textbf{83.17} & \textbf{90.81}
    & \textbf{70.87} & \textbf{74.09}
    & \cellcolor{red!6}\textbf{77.02}
    & \cellcolor{red!6}\textbf{82.45} \\
    & 1.5 & 83.05 & 90.74 & 70.74 & 73.96
    & \cellcolor{red!6}76.90 & \cellcolor{red!6}82.35 \\
    & 2.0 & 83.00 & 90.71 & 70.55 & 73.78
    & \cellcolor{red!6}76.77 & \cellcolor{red!6}82.25 \\
    \midrule
    \multirow{6}{*}{GF-7}
    & w/o Dir & 77.94 & 87.60 & 74.57 & 77.89
    & \cellcolor{red!6}76.25 & \cellcolor{red!6}82.75 \\
    & 0.2 & \textbf{78.52} & \textbf{87.97}
    & \textbf{75.73} & \textbf{79.03}
    & \cellcolor{red!6}\textbf{77.12}
    & \cellcolor{red!6}\textbf{83.50} \\
    & 0.5 & 78.48 & 87.94 & 75.70 & 78.99
    & \cellcolor{red!6}77.09 & \cellcolor{red!6}83.47 \\
    & 1.0 & 78.50 & 87.96 & 75.59 & 78.89
    & \cellcolor{red!6}77.05 & \cellcolor{red!6}83.43 \\
    & 1.5 & 78.41 & 87.90 & 75.38 & 78.71
    & \cellcolor{red!6}76.89 & \cellcolor{red!6}83.31 \\
    & 2.0 & 78.27 & 87.81 & 75.15 & 78.50
    & \cellcolor{red!6}76.71 & \cellcolor{red!6}83.16 \\
    \midrule
    \multirow{6}{*}{Planet}
    & w/o Dir & 45.71 & 62.74 & 42.64 & 44.25
    & \cellcolor{red!6}44.18 & \cellcolor{red!6}53.50 \\
    & 0.2 & 45.70 & 62.73 & 46.06 & 47.94
    & \cellcolor{red!6}45.88 & \cellcolor{red!6}55.34 \\
    & 0.5 & 45.44 & 62.48 & 46.10 & 48.02
    & \cellcolor{red!6}45.77 & \cellcolor{red!6}55.25 \\
    & 1.0 & 44.85 & 61.93 & 46.71 & 48.60
    & \cellcolor{red!6}45.78 & \cellcolor{red!6}55.27 \\
    & 1.5 & 44.84 & 61.92 & 46.31 & 48.20
    & \cellcolor{red!6}45.58 & \cellcolor{red!6}55.06 \\
    & 2.0 & \textbf{44.93} & \textbf{62.01}
    & \textbf{46.81} & \textbf{48.70}
    & \cellcolor{red!6}\textbf{45.87}
    & \cellcolor{red!6}\textbf{55.36} \\
    \midrule
    \multirow{6}{*}{ST-2}
    & w/o Dir & 28.21 & 44.00 & 23.18 & 23.83
    & \cellcolor{red!6}25.70 & \cellcolor{red!6}33.92 \\
    & 0.2 & 26.31 & 41.66 & 25.52 & 26.98
    & \cellcolor{red!6}25.92 & \cellcolor{red!6}34.32 \\
    & 0.5 & 26.17 & 41.48 & 27.76 & 29.61
    & \cellcolor{red!6}\textbf{26.97}
    & \cellcolor{red!6}\textbf{35.55} \\
    & 1.0 & 26.20 & 41.52 & 26.03 & 27.30
    & \cellcolor{red!6}26.11 & \cellcolor{red!6}34.41 \\
    & 1.5 & 25.30 & 40.39 & 26.45 & 27.69
    & \cellcolor{red!6}25.88 & \cellcolor{red!6}34.04 \\
    & 2.0 & 23.97 & 38.68
    & \textbf{28.00} & \textbf{29.96}
    & \cellcolor{red!6}25.99 & \cellcolor{red!6}34.32 \\
    \bottomrule
    \end{tabular}
    }
    \vspace{-0.0cm}
\end{table}

\subsubsection{Saddle-Aware Suppression}

\begin{table}[h]
    \centering
    \caption{Effect of saddle-aware penalty strength. ``w/o Sad'' denotes direction-aware regularization only. The saddle penalty is disabled for HR datasets with $\alpha_{\mathrm{HR}}=0$, while $\alpha_{\mathrm{LR}}$ controls the false-positive penalty in LR saddle regions.}
    \vspace{-0.2cm}
    \label{tab:sad-hyperparam}
    \scriptsize
    \setlength{\tabcolsep}{4pt}
    \renewcommand{\arraystretch}{0.95}
    \resizebox{0.484\textwidth}{!}{
        \begin{tabular}{lllcc|cc|cc}
        \toprule
        Dataset & $\alpha_{\mathrm{HR}}$ & $\alpha_{\mathrm{LR}}$
        & IoU & F1
        & B-IoU & B-F1
        & \cellcolor{red!10}M$_{\mathrm{IoU}}$
        & \cellcolor{red!10}M$_{\mathrm{F1}}$ \\
        \midrule
        \multirow{2}{*}{INRIA}
        & w/o Sad & -- & 83.13 & 90.79 & 70.54 & 73.79
        & \cellcolor{red!7}76.84 & \cellcolor{red!7}82.29 \\
        & 0 & -- & \textbf{83.29} & \textbf{90.88}
        & \textbf{70.86} & \textbf{74.09}
        & \cellcolor{red!7}\textbf{77.08}
        & \cellcolor{red!7}\textbf{82.49} \\
        \midrule
        \multirow{2}{*}{GF-7}
        & w/o Sad & -- & 78.48 & 87.94 & 75.70 & 78.99
        & \cellcolor{red!7}77.09 & \cellcolor{red!7}83.47 \\
        & 0 & -- & \textbf{78.68} & \textbf{88.07}
        & \textbf{75.76} & \textbf{79.04}
        & \cellcolor{red!7}\textbf{77.22}
        & \cellcolor{red!7}\textbf{83.56} \\
        \midrule
        \multirow{7}{*}{Planet}
        & -- & w/o Sad & 45.44 & 62.48 & 46.10 & 48.02
        & \cellcolor{red!6}45.77 & \cellcolor{red!6}55.25 \\
        & -- & 0 & \textbf{49.47} & \textbf{66.19}
        & 43.68 & 45.73
        & \cellcolor{red!7}46.57 & \cellcolor{red!7}55.96 \\
        & -- & 5 & 49.19 & 65.94 & 45.36 & 47.40
        & \cellcolor{red!7}47.28 & \cellcolor{red!7}56.67 \\
        & -- & 10 & 48.94 & 65.72 & 46.74 & 48.75
        & \cellcolor{red!7}\textbf{47.84}
        & \cellcolor{red!7}\textbf{57.24} \\
        & -- & 15 & 48.25 & 65.10 & 47.35 & 49.30
        & \cellcolor{red!7}47.80 & \cellcolor{red!7}57.20 \\
        & -- & 20 & 47.30 & 64.22
        & \textbf{48.23} & \textbf{50.13}
        & \cellcolor{red!7}47.77 & \cellcolor{red!7}57.18 \\
        & -- & 25 & 46.91 & 63.86 & 47.52 & 49.40
        & \cellcolor{red!7}47.21 & \cellcolor{red!7}56.63 \\
        \midrule
        \multirow{7}{*}{ST-2}
        & -- & w/o Sad & 26.17 & 41.48 & 27.76 & 29.61
        & \cellcolor{red!7}26.97 & \cellcolor{red!7}35.55 \\
        & -- & 0 & \textbf{37.97} & \textbf{55.04}
        & 28.25 & 29.44
        & \cellcolor{red!7}33.11 & \cellcolor{red!7}42.24 \\
        & -- & 5 & 37.48 & 54.52 & 29.25 & 30.46
        & \cellcolor{red!7}33.37 & \cellcolor{red!7}42.49 \\
        & -- & 10 & 36.88 & 53.89 & 29.41 & 30.59
        & \cellcolor{red!7}33.15 & \cellcolor{red!7}42.24 \\
        & -- & 15 & 36.73 & 53.72
        & \textbf{30.47} & \textbf{31.69}
        & \cellcolor{red!7}\textbf{33.60}
        & \cellcolor{red!7}\textbf{42.71} \\
        & -- & 20 & 36.22 & 53.17 & 29.88 & 31.11
        & \cellcolor{red!7}33.05 & \cellcolor{red!7}42.14 \\
        & -- & 25 & 35.57 & 52.47 & 30.40 & 31.64
        & \cellcolor{red!7}32.99 & \cellcolor{red!7}42.06 \\
        \bottomrule
        \end{tabular}
    }
    \vspace{-0.1cm}
\end{table}

\begin{figure}[h]
    \centering
    \includegraphics[width=0.98\linewidth]{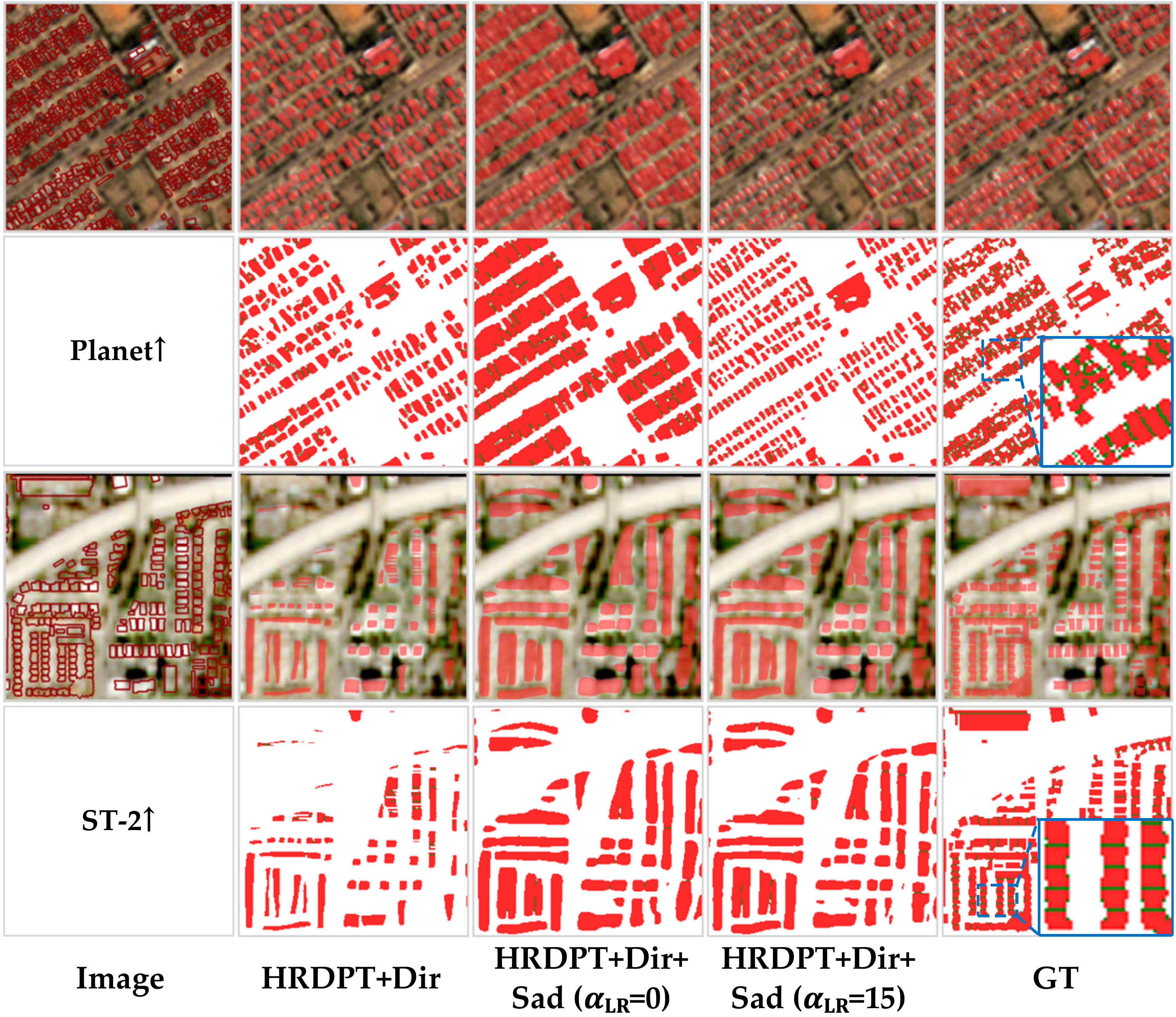}
    \caption{Visual comparison of saddle-weighted regularization, where \protect\textcolor[RGB]{0,150,0}{green} areas denote saddle regions. We compare CE+Dir, CE+Dir+Dice ($\alpha_{\mathrm{LR}}=0$), and CE+Dir+Saddle ($\alpha_{\mathrm{LR}}=15$), with all examples upsampled to 1\,m.}
    \label{fig:ablation_saddle}
    \vspace{-0.2cm}
\end{figure}

We then evaluate saddle-aware suppression, where $\alpha_{\mathrm{LR}}$ denotes the internal false-positive penalty strength. Since the saddle penalty is activated only for LR batches, its main effect appears on Planet and ST-2. As shown in Table~\ref{tab:sad-hyperparam}, $\alpha_{\mathrm{LR}}=15$ achieves the best overall performance, with M$_{\mathrm{IoU}}$ improving from 45.77 to 47.84 on Planet and from 26.97 to 33.60 on ST-2. The qualitative results in Fig.~\ref{fig:ablation_saddle} further show that saddle-aware suppression effectively preserves narrow gaps in LR imagery.

\subsection{Out-of-Domain Building Extraction Performance}
\label{sec:ood}

To assess OOD performance, we directly apply UniBuild to unseen domains without fine-tuning. The evaluated datasets are summarized in Table~\ref{tab:ood-datasets}. As shown in Table~\ref{tab:ood-unibuild}, UniBuild achieves strong zero-shot performance on Waterloo, WHU-Satellite, and Massachusetts, and remains effective on the more challenging ISPRS-Pforzheim dataset, demonstrating good transferability to unseen aerial and satellite imagery.

\begin{figure}[h]
    \centering
    \includegraphics[width=0.8\linewidth]{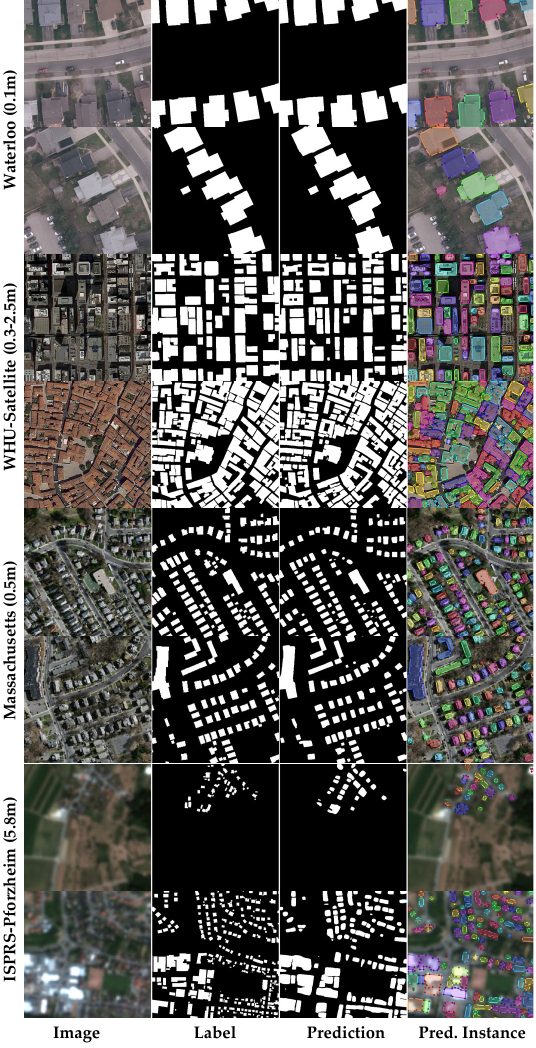}
    \vspace{-0.2cm}
    \caption{Qualitative OOD building extraction and polygonization examples of UniBuild on four unseen datasets. Each row corresponds to one OOD dataset, including Waterloo, WHU-Satellite, Massachusetts, and ISPRS-Pforzheim (bilinearly upsampled from 5.8 m to 1 m).}
    \label{fig:ood_visual_examples}
    \vspace{-0.3cm}
\end{figure}

Visual examples are provided in Fig.~\ref{fig:ood_visual_examples}. The results show that UniBuild produces spatially coherent masks with meaningful instance-level separability. We further apply a simple polygonization procedure, including connected-component separation, contour simplification, dominant-direction constraints, and short-edge merging, to visualize the extracted building instance structures. Overall, UniBuild directly produces reliable building masks and polygons, demonstrating practical robustness across diverse unseen domains. However, LR inputs may still limit accurate instance-level delineation when small or adjacent buildings are poorly resolved.

\begin{table}[h]
\centering
\caption{OOD datasets for zero-shot evaluation. For Massachusetts and ISPRS-Pforzheim, $0.5\ \mathrm{m}$ and $1.0\ \mathrm{m}$ denote the upsampled inference resolutions from $1.0\ \mathrm{m}$ and $5.8\ \mathrm{m}$, respectively.}
\vspace{-0.2cm}
\label{tab:ood-datasets}
\scriptsize
\setlength{\tabcolsep}{3.0pt}
\renewcommand{\arraystretch}{0.95}
\resizebox{0.49\textwidth}{!}{
\begin{tabular}{lllccc}
\toprule
Dataset & Region & Type & Subset & Resolution & Image size / Samples \\
\midrule
Waterloo \citep{he2022waterloo} & Canada & Aerial & Validation set & $0.12\,\mathrm{m}$ & $512 \times 512$ / 6,887 \\
WHU-Satellite \citep{ji2019multisource} & Global & Satellite & All & $0.3$--$2.5\,\mathrm{m}$ & $512 \times 512$ / 204 \\
Massachusetts \citep{mnih2013machine} & USA & Aerial & Test set & $1.0\,\mathrm{m}\,(0.5\,\mathrm{m}\uparrow)$ & $1500 \times 1500$ / 10 \\
ISPRS-Pforzheim \citep{isprs_zy3_stuttgart} & Germany & Satellite & One ZY-3 scene & $5.8\,\mathrm{m}\,(1.0\,\mathrm{m}\uparrow)$ & $42815 \times 14188$ / 1 \\
\bottomrule
\end{tabular}
}
\vspace{-0.3cm}
\end{table}

\begin{table}[h]
\centering
\caption{OOD evaluation results of UniBuild. ``B-IoU'' and ``B-F1'' denote Boundary-IoU and Boundary-F1, respectively. ``M$_{\mathrm{IoU}}$'' is the mean of IoU and B-IoU, while ``M$_{\mathrm{F1}}$'' is the mean of F1 and B-F1. Massachusetts-$0.5\,\mathrm{m}$ denotes the result after upsampling the original $1.0\,\mathrm{m}$ imagery by a factor of two.}
\vspace{-0.2cm}
\label{tab:ood-unibuild}
\scriptsize
\setlength{\tabcolsep}{6.0pt}
\renewcommand{\arraystretch}{1.0}
\resizebox{0.484\textwidth}{!}{
\begin{tabular}{lcccccc}
\toprule
Dataset & IoU & F1 & B-IoU & B-F1 & \cellcolor{red!10}M$_{\mathrm{IoU}}$ & \cellcolor{red!10}M$_{\mathrm{F1}}$ \\
\midrule
Waterloo & 89.28 & 94.34 & 73.15 & 76.86 & \cellcolor{red!7}81.22 & \cellcolor{red!7}85.60 \\
WHU-Satellite & 73.80 & 84.92 & 70.45 & 74.12 & \cellcolor{red!7}72.12 & \cellcolor{red!7}79.52 \\
Massachusetts-$1.0\,\mathrm{m}$ & 53.57 & 69.77 & 71.85 & 74.72 & \cellcolor{red!7}62.71 & \cellcolor{red!7}72.24 \\
Massachusetts-$0.5\,\mathrm{m}$ & 69.40 & 81.94 & 64.67 & 67.56 & \cellcolor{red!7}67.04 & \cellcolor{red!7}74.75 \\
ISPRS-Pforzheim & 32.51 & 49.07 & 30.77 & 32.63 & \cellcolor{red!7}31.64 & \cellcolor{red!7}40.85 \\
\midrule
Average & 63.71 & 76.01 & 62.18 & 65.17 & \cellcolor{red!7}62.95 & \cellcolor{red!7}70.59 \\
\bottomrule
\end{tabular}
}
\end{table}

The comparison between Massachusetts-$1.0\,\mathrm{m}$ and -$0.5\,\mathrm{m}$ highlights the scale-adaptive inference capability of UniBuild. By directly upsampling the input imagery from 1.0\,m to 0.5\,m before inference, UniBuild improves IoU from 53.57 to 69.40. This shows that UniBuild supports flexible inference-time resolution adjustment, allowing a practical trade-off between extraction accuracy and computational cost while adapting to different building scales without retraining.

\section{Conclusion}

In this paper, we presented \textbf{UniBuild}, a unified building extraction framework for multi-source optical RS imagery. UniBuild integrates multi-dataset training, an HR-DPT decoder, and geometry-aware regularization to improve transferable building representation learning, boundary recovery, and separation of adjacent buildings. Extensive experiments verify the effectiveness of UniBuild against competing models and show that its components are complementary, producing more accurate building regions, sharper boundaries, and clearer building separation. Multi-dataset training and out-of-domain evaluations further demonstrate its potential as a practical unified model for multi-source optical building extraction, including challenging imagery up to 10\,m GSD.

Despite these results, several limitations remain. First, building extraction from LR imagery, especially 10\,m Sentinel-2 imagery, remains challenging because small buildings and fine footprint details are close to or below the sensor resolution. Second, multi-source annotations may suffer from missing buildings, outdated footprints, and image--label misalignment. third, the current footprint regularization is used as post-processing rather than an end-to-end raster-to-vector component. Future work will focus on LR building modeling, noise-robust supervision, and end-to-end footprint vectorization.

\bibliographystyle{IEEEtran}
\bibliography{references}
\section{Supplemental Materials}

\subsection{Cross-Resolution Building Polygonization}

\begin{figure}[h]
    \centering
    \includegraphics[width=0.975\linewidth]{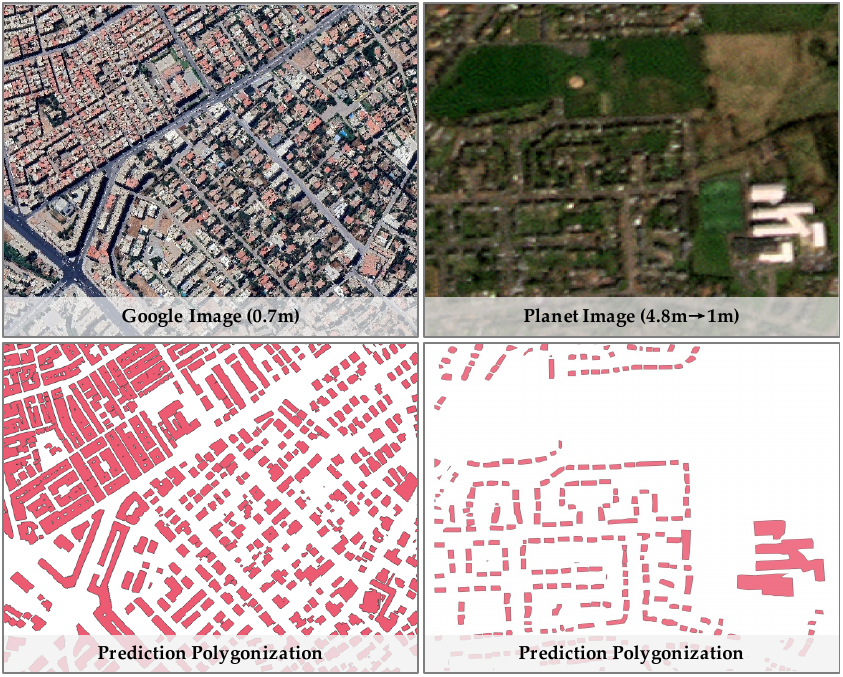}
    \caption{Building polygonization examples across image sources and spatial resolutions. The examples include Google Maps imagery from Fez, Morocco, at approximately 0.7\,m resolution, and Planet imagery from Carluke, Scotland, upsampled from approximately 4.8\,m to 1.0\,m before inference. From top to bottom, the rows show RGB images, UniBuild-derived building polygons, and corresponding OSM building polygons.}
    \label{fig:ood_google_crop}
\end{figure}

Fig.~\ref{fig:ood_google_crop} further demonstrates the practical image-to-map capability of UniBuild across different image sources and spatial resolutions. The examples include high-resolution Google Maps image crops from Fez, Morocco, with a spatial resolution of $0.7\ \mathrm{m}$, and Planet imagery from Carluke, Scotland, with an original spatial resolution of $4.8\ \mathrm{m}$, which is upsampled to $1.0\ \mathrm{m}$ before inference. Google Maps imagery represents an unseen data source, whereas Planet imagery is included in the training data but is evaluated here in a different geographic region. Given only RGB imagery, UniBuild can directly produce building masks, which are further converted into structured building polygons and exported as shapefiles using our simple and efficient polygonization algorithm provided in the released code. This enables an end-to-end workflow from newly acquired optical imagery to GIS-ready building vector data, making UniBuild useful for timely building map updating. The results also show that higher-resolution imagery leads to more detailed and accurate building polygons, while lower-resolution inputs are more suitable for large-scale building mapping than for fine-grained instance-level delineation. These results highlight the practical potential of UniBuild for GIS applications such as building map updating, urban monitoring, settlement mapping, and geospatial database enrichment.

\subsection{Zero-shot Building Change Detection}
\label{sec:change_detection}

Although the model is trained only for single-date building extraction, we further evaluate its zero-shot ability for building change detection on LEVIR-CD \citep{chen2020levircd} at a resolution of 0.5\,m. Given a bi-temporal image pair $(I_{t_1}, I_{t_2})$, we independently predict two-channel building probability maps $P_{t_1}$ and $P_{t_2}$ using UniBuild with a DINOv3-Base backbone trained on multiple datasets in Sec.~\ref{sec:multidataset-training}. The binary building masks $M_{t_1}$ and $M_{t_2}$ are obtained by pixel-wise argmax over the background and building output channels. The change map is then produced by applying an XOR operation to the two predicted masks:
\begin{equation}
C = M_{t_1} \oplus M_{t_2}.
\end{equation}
This protocol does not use any bi-temporal change labels during training, and therefore directly evaluates whether the learned building representation can support downstream temporal reasoning in a zero-shot manner.

\begin{figure}[h]
    \centering
    \includegraphics[width=1.0\linewidth]{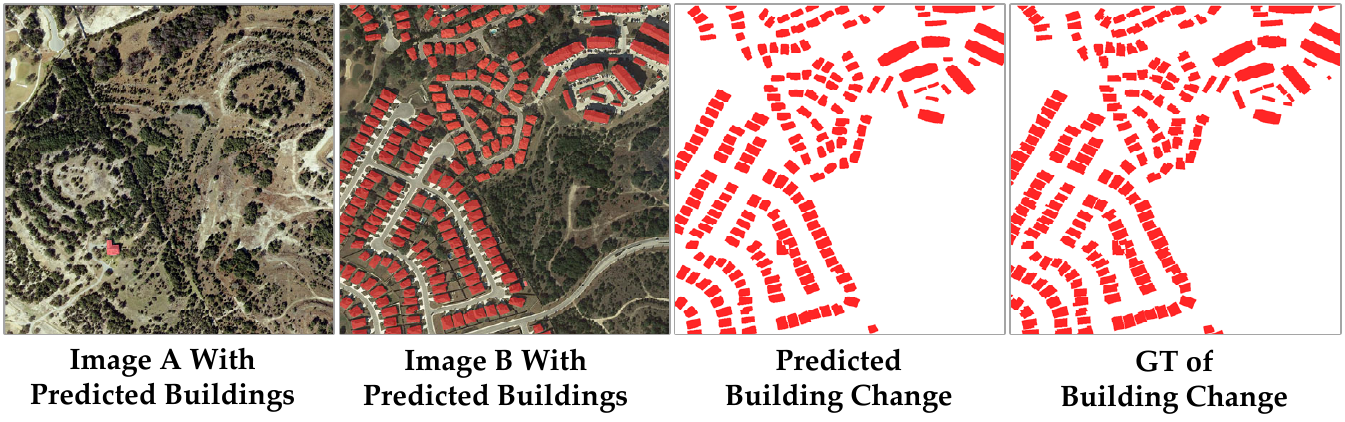}
    \caption{Zero-shot building change detection on LEVIR-CD.}
    \label{fig:ood_levir_cd}
\end{figure}

As shown in Table~\ref{tab:zero_shot_cd}, UniBuild achieves an F1 score of 85.14\% and an IoU of 74.12\% on LEVIR-CD without using any change-detection labels during training, with a qualitative samples shown in Fig.~\ref{fig:ood_levir_cd}. This result indicates that the learned building representation has strong zero-shot generalization ability and can be directly transferred from single-date building extraction to bi-temporal building change detection.

\begin{table}[h]
\centering
\caption{Zero-shot building change detection results on LEVIR-CD. The model is trained only for single-date building extraction and is not trained with bi-temporal change labels.}
\label{tab:zero_shot_cd}
\scriptsize
\setlength{\tabcolsep}{7pt}
\renewcommand{\arraystretch}{1.05}
\begin{tabular}{lcccc}
\toprule
Dataset & Precision & Recall & F1 & IoU \\
\midrule
LEVIR-CD & 88.26 & 82.23 & 85.14 & 74.12 \\
\bottomrule
\end{tabular}
\end{table}

\end{document}